\documentclass[a4paper,fleqn]{cas-sc}

\usepackage[authoryear,longnamesfirst]{natbib}

\usepackage{algorithmic}

\newcommand{\best}[1]{\textcolor{red}{\textbf{#1}}}
\newcommand{\second}[1]{\textcolor{blue}{\textbf{#1}}}

\begin{document}
\let\WriteBookmarks\relax
\def\floatpagepagefraction{1}
\def\textpagefraction{.001}

\shorttitle{LUMIN: Lifecycle-Utility-aware and Metadata-Informed Network}
\shortauthors{P. Yang \emph{et al.}}

\title[mode=title]{LUMIN: Lifecycle-Utility-aware and Metadata-Informed Network for Visual Anomaly Detection}

\author[1]{Pengfei Yang}
\cormark[1]
\ead{pfyangcrc@gmail.com}
\credit{}
\affiliation[1]{organization={Intelligent Precision Instrument},
            city={Shenzhen},
            country={China}}

\cortext[1]{Corresponding author}

\begin{abstract}
	Frozen-backbone memory-bank anomaly detectors pay their cost at three \emph{lifecycle}
	stages---cold-start bank construction, online retrieval, and offline metric evaluation---yet the
	sampling algorithms that build those banks spend most of their time on backbone forward passes,
	which is incompatible with industrial deployment. 
	
	To solve it, we propose LUMIN (Lifecycle-Utility-aware and Metadata-Informed Network), a visual anomaly detection
	framework. It comprises PICK (Pixel-metadata Informed Candidate Kit), a four-stage,
	plug-in-extensible sampling suite built on pixel \emph{metadata}, and two accuracy-neutral algorithms that accelerate
	network inference and performance evaluation, which together trading for speed without sacrificing accuracy and thereby make the framework \emph{utility-aware}. LUMIN therefore delivers a detector that is both
	fast (second-scale bank construction and inference on benchmarks like MVTec-AD) and accurate (matching
	state-of-the-art performance). 
	
	The artificial-intelligence contribution is the study of the detector. We examine how far the
	trainable parameters can be compressed, and find that 0.002M parameters---a single fused
	segmentation head---suffice for performance comparable to
	the state-of-the-art on five benchmarks.
\end{abstract}


\maketitle

\section{Introduction}\label{sec:introduction}

Industrial anomaly detection is a core component of quality control in intelligent manufacturing, aiming to automatically identify and localize surface defects~\cite{survey}. Unlike natural images, industrial deployment demands strict real-time performance ($<$100~ms per image) and low resource footprint.

Vision foundation models---CLIP~\cite{clip} and DINOv3~\cite{dinov3}---provide large-scale pre-trained representations, enabling methods such as WinCLIP~\cite{winclip}, AnomalyCLIP~\cite{anomalyclip}, VisualAD~\cite{visualad}, and UniADet~\cite{uniadet}. The dominant paradigm is a frozen backbone paired with a memory bank of normal patch features. Its cost is not paid at inference alone: it is spread across three life-cycle stages.

\begin{itemize}
	\item \textbf{Memory bank construction} (cold start, and again at every product switch). Memory-bank methods (e.g., PatchCore~\cite{patchcore}, UniADet~\cite{uniadet}) rely on nearest-neighbor retrieval, but their sampling algorithms---farthest-point sampling (FPS~\cite{fps}), K-means cluster-center sampling (KM~\cite{kmeans,kmeans++}), global statistical sampling (GSS) and greedy coreset~\cite{patchcore}---require numerous backbone forward passes and iterative distance computations, with construction times ranging from minutes to hours, failing rapid product-switching demands.

	\item \textbf{Online retrieval}: multi-scale features, cross-attention, or per-image serial similarity computation lead to low GPU utilization, high memory consumption, and poor batch scalability.

	\item \textbf{Offline metric evaluation} (development only): full-pixel evaluation on large-scale datasets such as Real-IAD~\cite{realiad} (129K images) dominates iteration cost.
\end{itemize}

All three stages are addressed here: PICK targets the first, and the two optimizations described below target the second and third. The first is the one specific to sample selection, and it raises two questions. \emph{Does sample selection actually buy accuracy on standard industrial benchmarks? And if it does not, along what axis should a sampler be chosen?}

We propose \textbf{LUMIN} (Lifecycle-Utility-aware and Metadata-Informed Network), a visual anomaly detection framework that answers both. Its sampling component, \textbf{PICK} (Pixel-metadata Informed Candidate Kit), is a four-stage, plug-in-extensible sampling suite built on pixel metadata; two further optimizations accelerate inference and performance evaluation at no cost in accuracy; and \textbf{FUSE} compresses the detector itself to a single fused segmentation head. The findings below introduce these in turn.

\begin{figure}
	\centering
	\includegraphics[width=\textwidth]{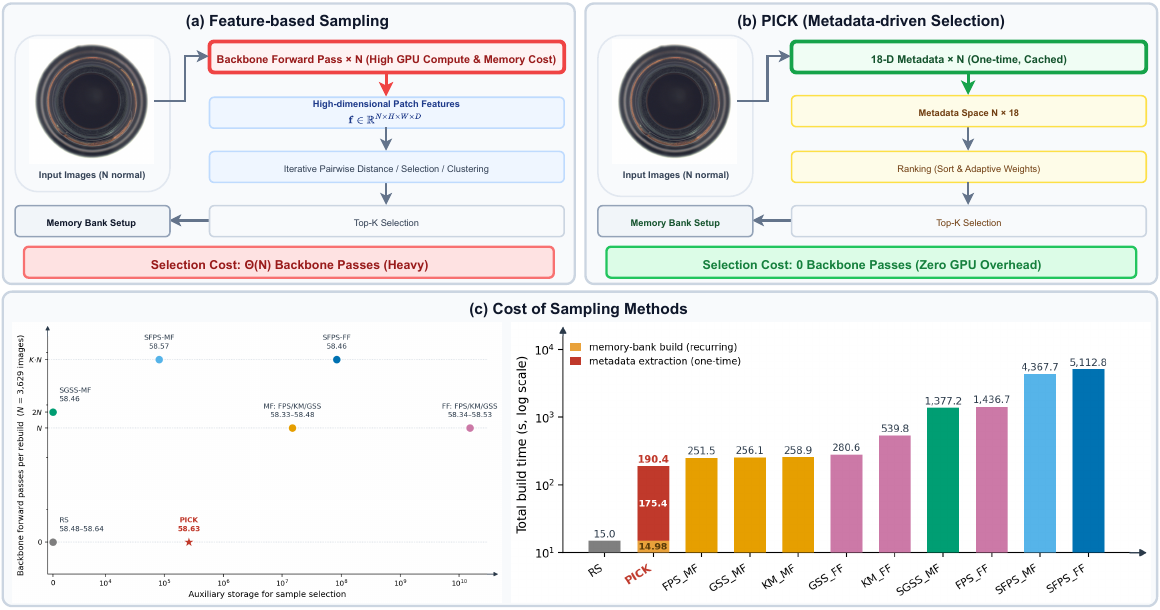}
	\caption{\textbf{Sampling a memory bank from cached metadata rather than from features.}
		(a)~Conventional feature-based sampling encodes every candidate before ranking
		($\Theta(N)$ backbone forward passes). (b)~PICK ranks a cached 18-D descriptor per image and
		selects with \emph{zero} backbone forward passes. (c)~Selection cost at $k{=}20$:
		structural cost on the left, measured build time on the right. The prefix \emph{S} marks
		streaming methods and the suffixes \_FF/\_MF full patch-feature maps and mean-pooled vectors.
		PICK's bar separates its one-time metadata extraction from a single rebuild, in unequal
		segments because the axis is logarithmic; every family's P-AUPR lies inside a band narrower
		than the seed noise.}
	\label{fig:overview}
\end{figure}

\textbf{Finding 1: PICK, and what its cost buys.} Bank construction is where the backbone time is spent, so PICK takes the backbone out of the sampling loop. PICK (Pixel-metadata Informed Candidate Kit), a four-stage adaptive memory bank sampling pipeline built on 18-dimensional pixel metadata and five complementary visual plugins. PICK completes sampling with zero backbone forward passes: metadata extraction is a one-time offline cost, coarse filtering is sub-second numerical sorting, and plugin layers are activated on demand. We nevertheless report the full pipeline throughout, and present Stages~3--4 as useful on a subset of scenes rather than a uniform source of gain. To report the cost honestly we decompose PICK\'s build into one-time offline preprocessing, recurring rebuild, and final bank assembly, and compare samplers on \emph{total} build time. Two accuracy-neutral inference and evaluation optimizations are reported as a separate engineering contribution. Fig.~\ref{fig:overview} summarises the contrast this section turns on: selection driven by backbone features versus selection driven by cached metadata (a, b), and the resulting cost landscape across sampler families (c).

PICK additionally supports progressive deployment, in two separable ways. First, the coarse-filter ratio is deliberately wide at cold start, which leaves Stages~3--4 a larger candidate pool to choose from; the ratio can be tightened once the line stabilizes. Second---and independent of the ratio---the memory bank is rebuilt only when the coarse ranking actually changes and a stable line pays nothing. In coarse-filter-only mode, where later stages are disabled, this admits a minimal deployment whose recurring cost is sub-second sorting.

\textbf{Finding 2: sampling saturates as the sample budget grows.} Sweeping three sampler
	families over five memory-bank seeds on MVTec-AD~\cite{mvtec}, we find that \emph{accuracy stops discriminating
	between samplers once the sample budget is large enough}: the spread between methods and the
	seed-to-seed variation of random selection both fall sharply with the sample budget, and beyond
	$k{=}10$ no pair of methods is separable (Table~\ref{tab:saturation}). Sample selection therefore
	matters only at very small budgets, and there it matters through the \emph{variance} of the outcome
	rather than its mean. Accuracy is not the axis on which a sampler should be chosen---which is why PICK is built around cost and determinism rather than around ranking quality.

\textbf{Finding 3: where frozen-backbone savings stop paying.} Two further costs are commonly assumed to follow from a frozen backbone. We quantify both, and report where they do not. (1) \emph{Cross-layer feature fusion before the detection head is only safe when backbone layers are representationally aligned.} On DINOv3, pre-detection single-head fusion reduces head parameters to $1/8$ of UniADet with minimal pixel-level loss; on CLIP, whose inter-layer heterogeneity is far larger, the same design fails in the zero-shot setting. (2) \emph{Head parameter reduction is not by itself an end-to-end lightweightness argument, because the frozen backbone dominates both memory and latency.} FUSE (Feature-fusion Unified Segmentation hEad) replaces $L$ per-layer segmentation heads with a single head; across the five benchmarks this moves pixel-level accuracy by a few points in either direction, with no consistent sign. 

The two modifications are interdependent by construction (Section~\ref{sec:lumin})---fusion produces the single representation that the single head then operates on.

\textbf{Scope.} LUMIN names the framework as a whole---a metadata-informed sampler, the two inference and evaluation optimizations, and the detector whose memory bank it is built for---and not any single module within it. We do not claim a new state-of-the-art detector, and we do not claim end-to-end lightweightness. The contributions below concern \emph{memory bank construction and evaluation cost}; FUSE's parameter reduction is reported as a single capacity contrast against the per-layer baseline, not as a deployment claim.

Systematic experiments cover five benchmarks (MVTec-AD~\cite{mvtec}, VisA~\cite{visa}, BTAD~\cite{btad}, KSDD~\cite{ksdd}, Real-IAD~\cite{realiad}), three detector architectures, and eleven sampling algorithms, offering reproducible baselines and practical guidelines for industrial deployment. The model and code will be publicly released upon acceptance.

\textbf{Contributions.}
\begin{itemize}
	\item \textbf{PICK}, a metadata-informed sampler that selects, with \emph{zero} backbone forward
	passes, images for building a memory bank; deterministic by construction, with ablations over hyper-parameters.
	\item Two accuracy-neutral engineering optimizations---parallel similarity retrieval and stratified
	evaluation---with their measured memory, latency, and evaluation-deviation characteristics.
	\item An open implementation of four sampler families---random selection, farthest-point sampling,
	K-means, and global statistical sampling---in their streaming and non-streaming, full-feature and
	mean-feature variants, benchmarked against PICK on latency, storage requirement, and
	pixel-level AUPR.
	\item A bounded analysis of pre-detection cross-layer fusion and detector head compression,
	including the negative CLIP result and its representation-level explanation.
\end{itemize}

\section{Related Work}\label{sec:related-work}

\subsection{Anomaly Detection Based on Vision Foundation Models}\label{sec:anomaly-detection}

CLIP~\cite{clip} enables zero-shot anomaly detection via vision-language alignment. WinCLIP~\cite{winclip} first achieves zero-shot localization with handcrafted prompts and multi-scale sliding windows---no learnable parameters, but costly due to per-window ViT re-encoding. AnomalyCLIP~\cite{anomalyclip} introduces learnable soft prompts and Diagonal Prominence Attention Maps, improving accuracy with 5.52M parameters, yet remains constrained by the full text encoder. Related zero- and few-shot variants refine the prompt and regularisation scheme~\cite{aprilgan,saa}, recast the task through large vision-language models that report anomalies in language~\cite{anomalygpt}, adapt prompts conditionally~\cite{adaclip}, or score unlabeled test images against one another~\cite{musc}; all of them inherit the text encoder that dominates their cost.

The paradigm this paper builds on predates foundation models: SPADE~\cite{spade} retrieves
patch-wise nearest neighbours from a bank of normal features, PaDiM~\cite{padim} models each patch
position as a multivariate Gaussian, and FastFlow~\cite{fastflow} fits a normalizing flow over frozen
features. What the foundation models change is the quality of the frozen features, not the retrieval
structure.

Recent work shifts from text supervision to purely visual paradigms. VisualAD~\cite{visualad} removes the text encoder entirely, inserting two learnable visual tokens into the patch sequence, reducing parameters by over 99\%. UniADet~\cite{uniadet} pushes this further with only $\sim$0.02M trainable parameters via decoupled classification-segmentation heads on frozen backbone features. We adopt UniADet as our baseline, first verifying the redundancy of its classification head, then exploring the extreme compression limit of the segmentation head.

\subsection{Memory Bank Sampling and Building Strategies}\label{sec:memory-bank}

PatchCore~\cite{patchcore} builds a memory bank of normal patch features for nearest-neighbor anomaly detection. Its greedy coreset sampling is effective but requires computing the full patch distance matrix $(O(N_{\text{patches}}^2)$), making it slow on large datasets. Alternatives---FPS~\cite{fps}, KM~\cite{kmeans,kmeans++}, global statistical sampling, and random selection---trade representativeness for speed. FPS and greedy coreset descend from the $k$-center and facility-location objectives~\cite{kcenter,facloc}, which bound how far any candidate lies from its nearest retained representative. Random selection offers no such bound but needs neither distances nor repeated passes, and its streaming form~\cite{reservoir} draws a uniform sample in one pass. Our PICK completes sampling with zero backbone passes; its coarse filtering is pure numerical sorting. PICK is orthogonal and complementary to PatchCore's patch-level coreset---PICK selects representative images while coreset compresses patch features, and the two can be combined to jointly optimize image coverage and storage efficiency.

\section{Method}\label{sec:method}

This section is organized along the logic of "sampling first, detection second." We begin with the two engineering contributions that make memory-bank construction affordable. The first is PICK (Section~\ref{sec:psp}), whose overall architecture, hierarchical process and core implementation principles are given below; its performance comparison against the other samplers is in Section~\ref{sec:psp-main-results}, its algorithmic advantages, mechanism analysis and deployment value are deferred to Section~\ref{sec:psp-analysis}, and its complexity classes to Appendix~\ref{sec:app-horizontal-comparison}. The second is the pair of optimizations that accelerate network inference and performance evaluation (Section~\ref{sec:optimization-strategies}). We then turn to the detector (Section~\ref{sec:lumin}), summarised in Fig.~\ref{fig:lumin}: panel~(a) shows the whole framework of LUMIN, and panel~(b) the three head configurations it passes through.

\subsection{Adaptive Memory Bank Sampling Pipeline Based on Metadata
and Multiple Plugins (PICK)}\label{sec:psp}

\subsubsection{Research Motivation and Analysis of Method
Advantages}\label{sec:motivation}

Memory bank construction is a core stage of feature-driven industrial anomaly detection: it builds a standard feature reference library by selecting high-quality normal samples and relies on nearest-neighbor retrieval for anomaly discrimination. Sample-selection quality directly determines detection accuracy, while construction efficiency directly constrains engineering deployability. Existing methods can be broadly categorized into streaming and non-streaming, both with inherent defects: non-streaming methods extract and cache high-dimensional features for the entire normal set at once, incurring substantial memory overhead; streaming methods alleviate global storage pressure but require repeated feature extraction, trading inference efficiency for storage optimization. Thus neither achieves the joint optimum of accuracy, computation, and storage, failing to meet the high-precision, low-latency, and high-stability demands of production lines.

To address these challenges, we leverage the prior characteristics of industrial scenes---standardized imaging, uniform backgrounds, and stable distributions---and move beyond the traditional iterative deep-feature selection paradigm to propose PICK (Pixel-metadata Informed Candidate Kit), a lightweight adaptive sampling pipeline. Relying on lightweight pixel-level metadata statistics and a multi-dimensional plugin adaptive fusion mechanism, PICK abandons the traditional iterative high-dimensional feature distance comparison paradigm and achieves high-precision, low-overhead, and strongly generalizable adaptive sampling without any loss of accuracy, fundamentally resolving the bottleneck where traditional algorithms cannot achieve both accuracy and efficiency.

\begin{figure}
	\centering
	\includegraphics[width=\linewidth]{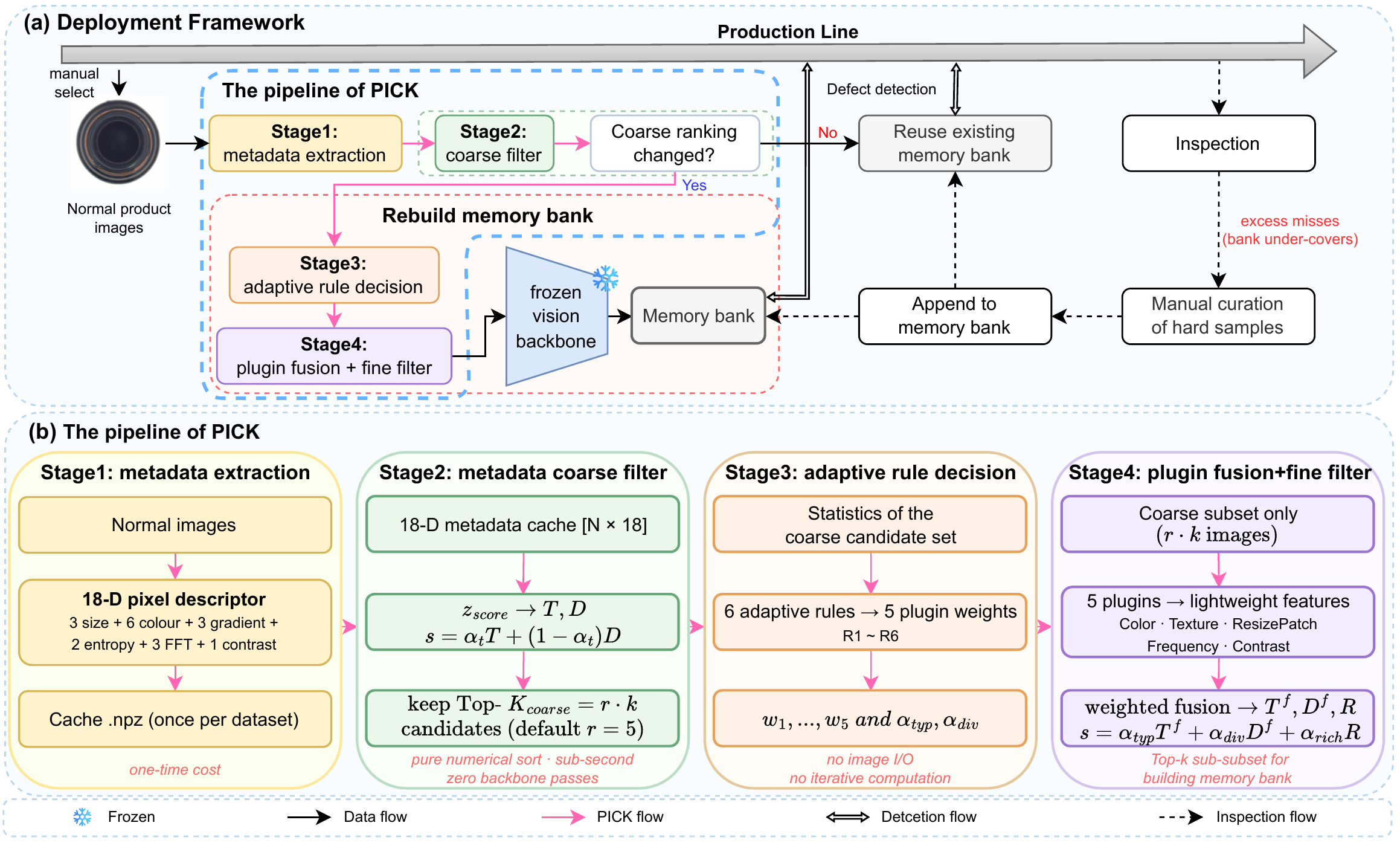}
	\caption{\textbf{The PICK pipeline and its deployment.}
		(a)~On a production line, each incoming batch is passed through Stages~1--2 and the bank is
		rebuilt only when the coarse ranking changes; detections the bank under-covers are curated
		manually and appended, closing the loop.
		(b)~The four stages---metadata extraction, coarse filtering, adaptive rule decision, and
		plugin fusion with fine filtering. Backbone features are computed after selection.
		Deployment is progressive in two respects. The coarse retention ratio is wide at cold start
		(default $r=5$) so that Stages~3--4 refine a broad candidate set, and narrows as the line
		stabilises; at $r=1$ they are skipped entirely and coarse ranking alone selects the bank. The
		manual-curation loop in (a) follows the same schedule.}
	\label{fig:psp}
\end{figure}

\subsubsection{Four-Stage Adaptive Sampling Pipeline}\label{sec:pipeline}

The PICK algorithm proposed in this paper adopts a four-stage
hierarchical pipeline architecture, as shown in Fig.~\ref{fig:psp}, following the layered optimization
design philosophy of "coarse filter first to reduce load, fine filter
later to improve quality, and dynamic weights to empower selection."
Throughout the entire process, sample filtering is completed via
lightweight CPU/GPU numerical operations without relying on real-time
backbone network inference. 

\textbf{Stage 1: Offline Metadata Extraction (One-Time Preprocessing).}
This stage replaces the traditional high-dimensional backbone feature extraction with an 18-dimensional pixel-level metadata descriptor. Formally, each normal image $I_i$ is mapped to a single descriptor vector
\begin{equation}
\begin{aligned}
	x_i = \big[\, & w,\; h,\; w/h,\;\; \mu_R,\mu_G,\mu_B,\;\; \sigma_R,\sigma_G,\sigma_B, \\
	              & \bar g,\; \sigma_g,\; g_{90},\;\; \mu_H,\; \sigma_H,\;\;
	                e_{\text{lo}},\, e_{\text{mid}},\, e_{\text{hi}},\;\; c_{\text{rms}} \,\big] \in \mathbb{R}^{18},
\end{aligned}
\label{eq:meta18}
\end{equation}
grouped into 3 geometry terms, 6 colour statistics, 3 Sobel gradient statistics, 2 $5\times5$ local-entropy statistics, 3 FFT band-energy ratios and 1 global-contrast term (Table~\ref{tab:metadata_dims}). Reading \eqref{eq:meta18} left to right: $w$ and $h$ are the image width and height in pixels, and $w/h$ is the aspect ratio; $\mu_c$ and $\sigma_c$ are the mean and standard deviation of colour channel $c\in\{R,G,B\}$ taken over all pixels of the image; $\bar g$, $\sigma_g$ and $g_{90}$ are the mean, standard deviation and 90th percentile of the Sobel gradient magnitude, characterising overall edge strength, its variability, and the strength of the strongest edges; $\mu_H$ and $\sigma_H$ are the mean and standard deviation of the $5\times5$ local entropy, a texture-complexity statistic that is largely invariant to colour; $e_{\text{lo}},e_{\text{mid}},e_{\text{hi}}$ are the fractions of FFT spectral energy falling in the low-, mid- and high-frequency bands, whose boundaries are fixed fractions of the radial spectrum; and $c_{\text{rms}}$ is the root-mean-square of the three RGB channel variances, a single global-contrast scalar. All 18 components are per-image scalars describing the image itself, independent of which candidate set the image later belongs to. Low-redundancy and high-stability, this descriptor captures global attributes and distributional properties of each sample. Extraction is a one-time operation per dataset: for a 512px input, single-image extraction takes $\sim$48~ms. All metadata is cached as a lightweight \texttt{.npz} file; for MVTec's 3,629 images, total cache size is $\sim$200~KB. Subsequent sampling iterations read the local cache directly, eliminating repeated feature extraction, raw-image I/O, and backbone forward passes---fundamentally resolving the efficiency bottleneck of traditional sampling algorithms.

\begin{table}
	\caption{Composition of the 18-dimensional (D) pixel metadata.}
	\label{tab:metadata_dims}
	\setlength{\tabcolsep}{2pt}
	\centering
	\scriptsize
	\begin{tabular*}{\tblwidth}{@{}LLL@{}}
		\toprule
		\textbf{Category} & \textbf{D} & \textbf{Description} \\
		\midrule
		Basic attributes & 3 & width, height, aspect ratio \\
		Color statistics & 6 & mean and standard deviation of RGB three channels \\
		Gradient magnitude & 3 & Sobel gradient mean, standard deviation, P90 quantile \\
		Local entropy & 2 & mean and standard deviation over 5$\times$5 window \\
		Frequency-domain \par energy & 3 & FFT low/mid/high-frequency energy ratios \\
		Global contrast & 1 & RMS mean of RGB channel variances \\
		\bottomrule
	\end{tabular*}
\end{table}

\textbf{Stage 2: Lightweight Metadata Coarse Filtering (Front-Loaded Candidate Reduction).} 
This stage uses the pre-cached 18-D metadata to perform rapid initial selection over the full normal sample set, with no backbone inference. Writing $\bar x,\hat\sigma\in\mathbb{R}^{18}$ for the per-dimension mean and standard deviation of the descriptor $x$ over the normal set (distinct from the per-image channel statistics $\mu_c,\sigma_c$ inside $x$ itself), each descriptor is standardised and its distance to the dataset centre is taken as a measure of how atypical the sample is:
\begin{equation}
	z_i = \frac{x_i-\bar x}{\hat\sigma+\epsilon}, \qquad d_i = \lVert z_i \rVert_2 ,
	\label{eq:coarse-z}
\end{equation}
where the division is element-wise and $\epsilon=10^{-6}$ prevents division by zero for a dimension that is constant across the normal set.
A typicality and a diversity term are then formed from that distance and combined linearly:
\begin{equation}
	T_i = \frac{1}{1+d_i}, \qquad
	D_i = \frac{d_i}{\max_j d_j + \epsilon}, \qquad
	s_i = \alpha_t\,\tilde T_i + (1-\alpha_t)\,\tilde D_i ,
	\label{eq:coarse-score}
\end{equation}
The two terms pull in opposite directions: $T_i$ rewards images close to the dataset centre, i.e.\ images that are \emph{typical} of normal production, while $D_i$ rewards images far from it, i.e.\ images that contribute \emph{coverage} of the normal variation. In \eqref{eq:coarse-score}, $\tilde{\cdot}$ denotes min--max normalisation to $[0,1]$ over the normal set (applied separately to $T$ and $D$, since $T_i\in(0,1]$ while $D_i$ is unbounded above), $\alpha_t$ is a fixed blend weight (\texttt{coarse\_alpha\_typical}, default 0.5), and $s_i$ is the resulting global typicality score. Given a target memory bank of $K$ images, the stage retains $K_{\text{coarse}}=\lceil rK\rceil$ candidates with the largest $s_i$, where $r$ is the coarse retention ratio (\texttt{coarse\_keep\_ratio}, default 5); for 20-shot, this retains the top 100 representative samples, substantially reducing subsequent fine-grained computation. Total execution time is under 1~s. Since \eqref{eq:coarse-score} is a closed-form function of cached scalars, this module can independently support the complete sampling process when \texttt{keep\_ratio=1.0}, serving as a near-zero-overhead minimal deployment scheme that fits standardized industrial scenarios with stable imaging and data distributions.

\begin{table}
	\caption{Adaptive decision rules for dynamic plugin weighting ($\uparrow$ denotes increasing the corresponding plugin weight).}
	\label{tab:adaptive_rules}
	\setlength{\tabcolsep}{3pt}
	\centering
	\scriptsize
	\begin{tabular*}{\tblwidth}{@{}LLL@{}}
		\toprule
		\textbf{Rule} & \textbf{If} & \textbf{Then} \\
		\midrule
		R1 & High RGB variance & $\uparrow$ ColorPlugin \\
		R2 & High gradient variance & $\uparrow$ TexturePlugin (HOG) \\
		R3 & High local entropy variance & $\uparrow$ TexturePlugin (LBP) \\
		R4 & High scale/contrast variance & $\uparrow$ ResizePatchPlugin \\
		R5 & High high-frequency energy & $\uparrow$ FrequencyPlugin \\
		R6 & High brightness dynamic range & $\uparrow$ ContrastPlugin \\
		\bottomrule
	\end{tabular*}
\end{table}

\textbf{Stage 3: Adaptive Rule Decision (Dynamic Weight Adaptation).}
The six rules and their mapping to plugin weights are listed in Table~\ref{tab:adaptive_rules}. This stage is an adaptive rule-decision module. Based on global distribution characteristics of the coarse-filtered candidates, it dynamically solves plugin fusion weights and fine-grained thresholds via multi-dimensional customized rules. This involves only lightweight numerical statistics with no iterative computation or I/O overhead. It adaptively allocates each plugin's contribution and optimizes evaluation criteria per scene, overcoming the fixed-weight limitations and weak generalization of traditional methods, and provides the scene-optimal scoring strategy for subsequent filtering.

Specifically, let $V_d$ be the variance of metadata dimension $d$ computed over the coarse candidate set. Each of the six rules contributes additively to the raw relevance $u_j$ of one plugin, starting from a floor of $0.1$:
\begin{equation}
\begin{aligned}
	u_{\text{Col}} &= 0.1 + \!\!\sum_{d \in \{\sigma_R,\sigma_G,\sigma_B\}}\!\! V_d, \\[2pt]
	u_{\text{Tex}} &= 0.1 + 1.5\!\!\sum_{d \in \{\bar g,\sigma_g,g_{90}\}}\!\! V_d
	                  + 1.5\!\!\sum_{d \in \{\mu_H,\sigma_H\}}\!\! V_d, \\[2pt]
	u_{\text{Res}} &= 0.1 + 0.8\,(V_{\text{ar}} + V_{c}), \\[2pt]
	u_{\text{Fre}} &= 0.1 + \!\!\sum_{d \in \{e_{\text{lo}},e_{\text{mid}},e_{\text{hi}}\}}\!\! V_d
	                  + 2.0\,\overline{e_{\text{mid}}}, \\[2pt]
	u_{\text{Con}} &= 0.1 + V_{c} + 0.5\!\!\sum_{d \in \{\mu_R,\mu_G,\mu_B\}}\!\! V_d,
\end{aligned}
\label{eq:rules}
\end{equation}
where the subscripts abbreviate ColorPlugin, TexturePlugin, ResizePatchPlugin, FrequencyPlugin and ContrastPlugin; $V_{\text{ar}}$ and $V_c$ are the variances of the aspect ratio and of the global-contrast term; and $\overline{e_{\text{mid}}}$ is the mean mid-band FFT energy. The rules thus fire on \emph{dispersion} rather than on absolute level: a scene is deemed to need a plugin when the metadata dimensions that plugin targets are themselves variable across the normal set. The raw scores are then normalised and blended with a per-category prior $p$:
\begin{equation}
	w = \frac{\alpha_p\,p + (1-\alpha_p)\,\mathrm{softmax}(u)}
	         {\sum_{j} \big[\alpha_p\,p + (1-\alpha_p)\,\mathrm{softmax}(u)\big]_j},
	\label{eq:weights}
\end{equation}
with prior strength $\alpha_p$ (\texttt{prior\_strength}, default 0.3) and $p$ a five-vector supplied by a scene configuration file (a uniform default is used when absent), so that domain knowledge can be injected without retraining. Finally, the module computes the typicality and diversity coefficients used in Stage~4, from the total dispersion $V=\sum_d V_d$ of the candidate set:
\begin{equation}
\begin{aligned}
	\alpha_{\text{typ}} &= \mathrm{clip}\!\left(0.6 - 0.4\tanh \tfrac{V}{5000},\; 0.2,\; 0.6\right), \\[2pt]
	\alpha_{\text{div}} &= \mathrm{clip}\!\left(0.2 + 0.4\tanh \tfrac{V}{5000},\; 0.1,\; 0.5\right).
\end{aligned}
\label{eq:alphas}
\end{equation}
Both are bounded away from the extremes, so that neither term can dominate under a single unusual scene; a tightly clustered normal set (small $V$) shifts weight toward typicality, a dispersed one toward diversity.

\textbf{Stage 4: Multi-Plugin Feature Fusion and Fine-Grained Filtering.}
Stage~4 targets the coarse-filtered candidate set and activates five lightweight plugins to extract fine-grained features across color, texture, spatial layout, frequency-domain periodicity, and lighting contrast. The weighted plugin outputs are concatenated and the result re-normalised,
\begin{equation}
	f_i = \frac{\big\|_{j:\,w_j>0}\; w_j\,\varphi_j(I_i)}
	           {\big\|\,w_j\,\varphi_j(I_i)\,\big\|_2},
	\label{eq:fusefeat}
\end{equation}
where $\varphi_j(I_i)$ is the feature vector that plugin $j$ extracts from image $I_i$ (its dimension differs per plugin), $I_i$ is the $i$-th surviving coarse candidate, and $\|$ denotes concatenation. Plugins whose adaptive weight $w_j$ is negligible are never evaluated at all, which is where much of Stage~4's cost saving comes from.

Let $N_c\le\lceil rK\rceil$ be the number of candidates that survived coarse filtering and $F=[f_1,\dots,f_{N_c}]^{\top}$ the matrix of their fused features. Three complementary scores are computed on $F$:
\begin{equation}
	D^{\text{f}}_i = \lVert f_i - \bar f \rVert_2, \quad
	T^{\text{f}}_i = \frac{1}{\lVert f_i - c_{k(i)} \rVert_2 + \epsilon}, \quad
	R_i = \mathrm{Var}_d\big(f_{i,d}\big),
	\label{eq:three-scores}
\end{equation}
where $\bar f$ is the mean of the fused features over the candidate set, $c_{k(i)}$ is the centre of the MiniBatchKMeans cluster to which $f_i$ is assigned ($c=\min\!\big(8,\max(2,\lfloor N_c/5\rfloor)\big)$ clusters, so that the partition adapts to the pool size), and $R_i$ is the variance of $f_i$ across its own dimensions, measuring how broadly a sample excites the fused representation. The superscript $\text{f}$ marks that $D^{\text{f}}_i$ and $T^{\text{f}}_i$ are the dispersion and cluster-typicality analogues of the Stage-2 terms, but computed in the plugin-fused space rather than on 18-D metadata; the two are distinct quantities that happen to share a name, and are normalised over the candidate set just as in \eqref{eq:coarse-score}. After min--max normalisation of each term, the composite score is
\begin{equation}
	\mathrm{score}_i = \alpha_{\text{typ}}\tilde T^{\text{f}}_i + \alpha_{\text{div}}\tilde D^{\text{f}}_i + \alpha_{\text{rich}}\tilde R_i,
	\qquad \alpha_{\text{rich}} = 1 - \alpha_{\text{typ}} - \alpha_{\text{div}},
	\label{eq:composite}
\end{equation}
and the $K$ highest-scoring candidates form the memory bank. \emph{Note that Stages~2 and~4 both rank by a typicality--diversity composite, but over different representations and with different roles}: Stage~2 scores the 18-D metadata over the \emph{full} normal set purely to shrink the candidate pool at zero backbone cost, whereas Stage~4 scores the plugin-fused features over the \emph{reduced} pool to choose the final $K$, and adds a richness term that no individual metadata statistic expresses. The plugin architecture is plug-and-play, supporting flexible activation, deactivation, and custom extension, adapting to diverse industrial scenarios with complex textures, lighting fluctuations, and scale variations.

The five plugins are as follows:
\begin{itemize}
	\item \textbf{ColorPlugin}: RGB/HSV histograms and per-channel statistics for background color, deviation, and shift.
	\item \textbf{TexturePlugin}: HOG (gradient/structure) and LBP (local texture density) for wear, scratches, wrinkles, and texture disarray.
	\item \textbf{FrequencyPlugin}: FFT-based high/low-frequency energy and periodic-texture responses for subtle periodic anomalies.
	\item \textbf{ContrastPlugin}: global brightness range, local contrast, and lighting-gradient distribution for highlights, reflections, shadows, and exposure anomalies.
	\item \textbf{ResizePatchPlugin}: multi-scale patch sampling for spatial layout, position shift, and structural disarray.
\end{itemize}

After extraction, each plugin branch is globally Min-Max normalized, then weighted fused with Stage~3 weights. Candidates are re-ranked by typicality, diversity, and feature richness, and Top-$K$ samples are selected to build a lightweight, robust memory bank.

\subsection{Optimization Strategies for Model Inference and
Evaluation}\label{sec:optimization-strategies}

To further improve model inference efficiency, reduce GPU memory
overhead, and accelerate experiment iteration, we design two
optimization strategies---parallel similarity computation and stratified
sampling evaluation---targeting the two bottleneck stages of model
matching inference and pixel-level metric evaluation. Neither
optimization changes the model's core training and
prediction logic or incurs any loss of detection accuracy, effectively
improving the model's engineering deployment performance
and experiment iteration efficiency.

\subsubsection{Parallel Computation Optimization for Memory Bank
Retrieval}\label{sec:parallel-retrieval}

During few-shot matching, the model computes cosine similarity between query patches and memory prototypes to determine anomaly scores. A naive on-the-fly implementation of \texttt{F.cosine\_similarity} on a per-image, per-dimension basis would materialize a $[C, M, D]$ intermediate tensor ($C$: query patches, $M$: memory bank size, $D$: feature dimension), incurring a massive memory footprint and frequently triggering OOM errors under high-resolution inputs, crippling large-scale inference. Worse, its serial per-sample traversal underutilizes GPU parallelism, resulting in high latency and low throughput that undermine real-time performance and deployability.

To resolve this bottleneck, we leverage the equivalence $\cos(\boldsymbol{q},\boldsymbol{m})=\frac{\boldsymbol{q}\cdot\boldsymbol{m}}{\|\boldsymbol{q}\|\|\boldsymbol{m}\|}= \left(\frac{\boldsymbol{q}}{\|\boldsymbol{q}\|}\right)\cdot\left(\frac{\boldsymbol{m}}{\|\boldsymbol{m}\|}\right)$. Our optimization operates in two steps: (1) pre-normalize and cache all memory prototypes offline; (2) during inference, normalize only the query features and compute the full similarity matrix via $\boldsymbol{Q}_{norm}\boldsymbol{M}_{norm}^\top$. This eliminates the 3D intermediate tensor, reducing memory complexity from $O(C\cdot M\cdot D)$ to $O(C\cdot M)$.

Matrix multiplication enables fully parallel matching, exploiting GPU parallelism and keeping latency nearly flat as batch size increases, with no additional overhead. This is the exact (non-approximate) counterpart of the GPU similarity search that libraries such as FAISS~\cite{faiss} implement for banks too large to scan exhaustively; here $M$ is the $k$-shot bank size, small enough that exact search is the right choice. Experiments confirm numerically identical similarity values and unchanged P-AUPR: latency and memory both fall dramatically, without cost in accuracy, as shown in Table~\ref{tab:retrieval_chunk}.

\subsubsection{Stratified Sampling for Pixel-Level Evaluation}\label{sec:stratified-sampling}

Pixel-level evaluation metrics (P-AUROC, P-AUPR, P-F1max) rely on prediction scores and ground-truth labels of all pixels. Large-scale datasets like Real-IAD~\cite{realiad} contain massive images; full-pixel evaluation generates heavy floating-point operations, greatly slowing iteration and increasing hardware cost---a major bottleneck for model tuning and ablation studies.

To balance accuracy and iteration efficiency, we introduce a stratified sampling strategy applied exclusively at the evaluation stage. It does not participate in forward propagation or prediction, nor alter predictions or feature distributions; its sole effect is reducing computational overhead.

For each category separately, let $N$ be the pixel count pooled over that category's test images, with $N_{\text{pos}}$ anomalous and $N_{\text{neg}}$ normal pixels. Given a sampling ratio, the theoretical sampled count is $\tilde{N}_{\text{samp}} = N \cdot \text{ratio}$. The purpose of the floor is to stop paying statistical error for a saving that no longer exists: the metric call has a fixed cost that dominates once the sample is small, so below a certain budget the ratio is no longer a useful knob (Section~\ref{sec:metric-acceleration} measures this). We therefore set a per-category lower bound $N_{\text{lower}}=100000$, and the effective sampling size is $N_{\text{samp}} = \max(\lceil \tilde{N}_{\text{samp}} \rceil, N_{\text{lower}})$.

If $N \le N_{\text{samp}}$, the full pixel set is used directly, preserving high-precision evaluation. Otherwise, positive and negative sample counts are allocated proportionally:
\begin{align}
	N_{\text{pos}}^s &= \min\left(\left\lfloor N_{\text{samp}} \cdot \frac{N_{\text{pos}}}{N} \right\rfloor,\, N_{\text{pos}}\right) \\
	N_{\text{neg}}^s &= \min\left(N_{\text{samp}} - N_{\text{pos}}^s,\, N_{\text{neg}}\right)
\end{align}

Sampling is then performed separately on the anomaly and normal pixel sets, and the two subsets are merged for metric computation.

Both subsets are drawn without replacement and merged; the positive count is truncated (rather than rounded) from the proportional allocation, the negative count takes the remainder, both are capped at their pool size, and a budget that allocates zero positive pixels is not evaluated at all.

This strategy strictly preserves the original positive/negative distribution, suppressing statistical error from random sampling. It substantially reduces evaluation cost, accelerating iteration and ablation validation; the residual metric deviation is bounded and reported in Section~\ref{sec:metric-acceleration}: at or above the floor it does not exceed 0.064 P-AUPR over ten evaluation seeds.

\subsection{From Decoupled Branches to a Fused Single Head}\label{sec:lumin}

\begin{figure}
	\centering
	\centerline{\includegraphics[width=\linewidth]{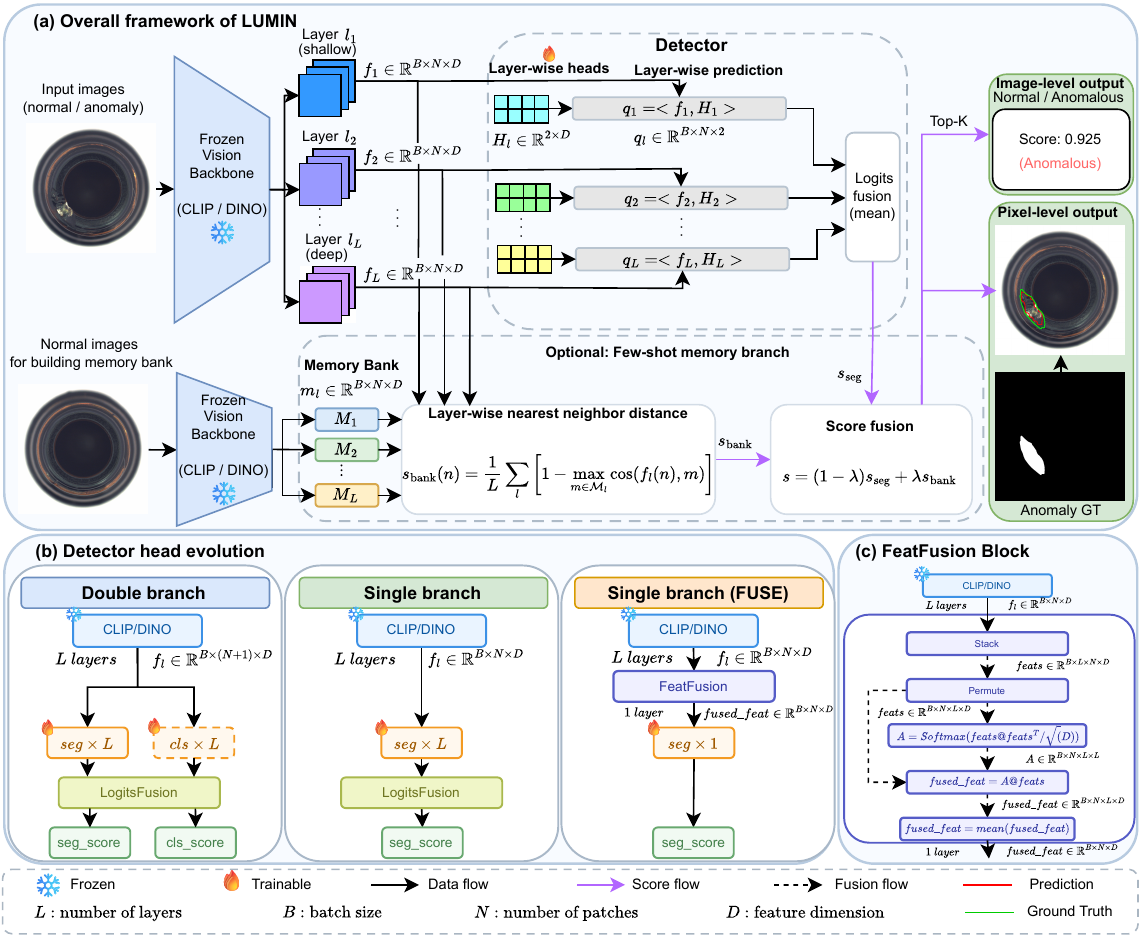}}
	\caption{\textbf{The LUMIN detector.}
	(a)~The overall framework. A frozen backbone supplies features from $L$ layers; each layer is
	scored against its own head and, when a memory bank is available, against that layer's nearest
	neighbours, the two scores being combined into $s$. The image-level verdict is the Top-$K$ mean of
	the pixel-level scores, and the optional few-shot memory branch is drawn dashed.
	(b)~Detector head evolution, the subject of Section~\ref{sec:lumin}: the Double branch feeds $N$
	patch features plus a class token into separate segmentation and classification heads, the class
	token carrying the image-level score; the Single branch drops it and recovers the score from the
	pixel map; FUSE fuses the layers first, leaving a single segmentation head.
	(c)~FeatFusion, the parameter-free block that forms and applies the inter-layer attention $A$ of
	Eq.~\eqref{eq:layer-attention}.}
	\label{fig:lumin}
\end{figure}

Our foundation is the \textbf{Double branch} design of UniADet~\cite{uniadet}
(Fig.~\ref{fig:lumin}(b), left): the frozen backbone feeds $N$ patch tokens \emph{and} one class token
per image, $f_i\in\mathbb{R}^{B\times(N+1)\times D}$, into two independently trained heads---a
segmentation head producing the pixel-level anomaly map, and a classification head reading the class
token for the image-level score. A Bayesian search (Optuna~\cite{optuna}) puts the optimal
classification temperature more than an order of magnitude above the segmentation temperature, so the
two branches plainly want separate tuning.

\textbf{Single branch (UniADet\_seg).} Removing the classification branch, its class token and its
optimizer halves the trainable parameters ($4LD\to 2LD$) and proportionally cuts training memory and
latency. Image-level scores are then recovered as the Top-$K$ mean of the pixel map, requiring no
further training. The price is structural rather than a design flaw: image-level discrimination now
inherits every error of pixel localization, a one-way pixel-to-image flow.

What remains exposed is the head itself---$L$ independent sets of per-layer weights, an overfitting
risk on a small industrial dataset. FUSE addresses this with two modifications that are interdependent
by construction, because fusion produces the single representation the shared head then operates on.

\textbf{Cross-layer feature fusion (FeatFusion).} UniADet\_seg feeds each layer to its own head and
fuses only at the logits. We move fusion ahead of detection: the $L$ layer features are stacked into
$F\in\mathbb{R}^{L\times B\times N\times D}$ and passed through a parameter-free cross-layer
self-attention~\cite{attention}. Writing $F_{b,n}\in\mathbb{R}^{L\times D}$ for the layers at image $b$
and patch $n$,
\begin{equation}
	A_{b,n} = \mathrm{softmax}\left(\frac{F_{b,n}\,F_{b,n}^{\top}}{\sqrt{D}}\right) \in \mathbb{R}^{L\times L},
	\qquad F_{\text{fused}}[b,n] = A_{b,n}\,F_{b,n}.
	\label{eq:layer-attention}
\end{equation}
The attention is computed independently at every patch, so which layers carry the most useful evidence
can vary spatially, at no parameter cost.

\textbf{Segmentation-head compression.} The fused representation no longer needs per-layer weights, so
we replace the $L$ sets (each $\mathbb{R}^{2\times D}$) with a single
$W_{\text{seg}}\in\mathbb{R}^{2\times D}$: $1/L$ of the parameters, shrinking the model from
$\sim$37KB to $\sim$11KB at $L=4$, with the advantage growing in $L$. Table~\ref{tab:architecture_evolution} compares the three configurations.

\begin{table}
	\caption{Architecture evolution: comparison of UniADet, UniADet\_seg, and FUSE with $L=4$}
	\label{tab:architecture_evolution}
	\setlength{\tabcolsep}{3pt}
	\centering
	\scriptsize
	\begin{tabular*}{\tblwidth}{@{}LLLL@{}}
		\toprule
		Component & Double Branches & Single Branch & FUSE \\
		\midrule
		Fusion & (Post-) Scores & (Post-) Scores & (Pre-) Features \\
		Segmentation Head & Multiple & Multiple & Single \\
		Trainable Parameters & $\sim$16K ($4LD$) & $\sim$8K ($2LD$) & $\sim$2K ($2D$) \\
		Size of .pth File  & $\sim$72KB & $\sim$37KB & $\sim$11KB \\
		\bottomrule
	\end{tabular*}
\end{table}

\section{Experiments}\label{sec:experiments}

\subsection{Experimental Setup}\label{sec:arch-experimental-setup}

We evaluate on five public industrial anomaly detection benchmarks covering diverse scenarios:
MVTec-AD~\cite{mvtec}, VisA~\cite{visa}, BTAD~\cite{btad}, KSDD~\cite{ksdd} and
Real-IAD~\cite{realiad}. 

\textbf{Sampling.} We compare PICK against four baseline families---random selection (RS),
farthest-point sampling (FPS), K-means cluster-center sampling (KM) and global statistical sampling (GSS)---over
$k\in\{0,1,4,10,20\}$; metrics are P-AUPR and construction time. A leading \emph{S} marks a streaming
variant, and the suffixes \_FF/\_MF denote full patch-feature maps and mean-pooled 1-D vectors. The
baselines sample with DINOv3 ViT-L/16~\cite{vit,dinov3}, taking features from backbone layer~12. We fix
the seed for inference and metric computation and sweep it for sampling and memory-bank construction,
logging the two separately so that selection variance is not confused with other noise. All configurable
hyperparameters, their defaults and their functional roles are listed in
Table~\ref{tab:psp_hyperparam_overview}; every experiment adopts them without further random tuning.

\textbf{Detector.} Models are trained on VisA and tested directly on the other four without fine-tuning,
and VisA is evaluated with a model trained on MVTec-AD; we report image- and pixel-level AUROC, AUPR and
F1-max. Frozen CLIP and DINOv3 backbones at $518\times518$ and $512\times512$, features from layers
$\{6,12,18,24\}$, batch size~8 and learning rate $10^{-3}$ are shared by every configuration except
AnomalyCLIP~\cite{anomalyclip}, which we run in its reference form---the official implementation, its
released VisA-trained checkpoint, and a \emph{single} backbone layer $\{24\}$ (one-indexed) feeding the
segmentation branch. That exception is not a tuning choice: under exactly the protocol of
Table~\ref{tab:0shot}, widening AnomalyCLIP's feature set to $\{6,12,18,24\}$ lowers its pixel-level
AUROC from 80.6 to 71.1. Baselines are AnomalyCLIP, VisualAD~\cite{visualad} and UniADet~\cite{uniadet},
alongside our FUSE. All experiments run on a single 48G NVIDIA L40S GPU.

\subsection{Sampler Comparison: Saturated Accuracy, Cost and Determinism}\label{sec:psp-main-results}\label{sec:cost-model}

The full eleven-sampler sweep is deferred to Appendix~\ref{sec:app-horizontal-comparison}, where Table~\ref{tab:sampling_results_compare} reports P-AUPR against build time at $k\in\{0,1,4,10,20\}$ for every family and Table~\ref{tab:sampling_theory_compare} gives the complexity classes behind them. What the rest of the paper rests on is the seed-controlled comparison of Table~\ref{tab:saturation}, read below.

\textbf{Accuracy does not separate the samplers.} Over five memory-bank seeds and four shot settings (Table~\ref{tab:saturation}), PICK, random selection and $K$-means sit in the same narrow band: the spread between their means never exceeds 0.37 P-AUPR and falls to 0.06 at $k{=}20$, comparable to the per-category variability (paired $\sigma=0.35$--$1.06$), and no pairwise comparison across the 15 categories reaches significance ($p\ge0.058$). At $k{=}1$ the ordering is decided by the seed variance of the randomized methods rather than by how well they choose: random selection carries $\sigma=0.204$ there, an order of magnitude above its spread at $k\ge4$, while PICK's value is fixed at every $k$. Accuracy being saturated, the axes left to compare are construction cost, resource occupancy and determinism---the contrast Fig.~\ref{fig:overview} is built around, and the one this paper adopts.

\begin{table}
	\caption{Saturation of memory-bank sampling on MVTec-AD. P-AUPR is the mean over the 15 categories and five memory-bank seeds, with a fixed evaluation seed, so that only the composition of the bank varies. The $\sigma$ columns are the standard deviation of each method across the five memory-bank seeds. The final column is the spread between the three method means.}
	\label{tab:saturation}
	\setlength{\tabcolsep}{3pt}
	\centering
	\scriptsize
	\begin{tabular*}{\tblwidth}{@{}LLLLLLLL@{}}
		\toprule
		$k$ & PICK & Random & $K$-means & $\sigma_{\text{PICK}}$ & $\sigma_{\text{rand}}$ & $\sigma_{\text{KM}}$ & method spread \\
		\midrule
		1  & 57.22 & 57.18 & 57.55 & 0.00 & 0.20 & 0.00 & 0.37 \\
		4  & 58.26 & 58.37 & 58.44 & 0.00 & 0.04 & 0.03 & 0.19 \\
		10 & 58.71 & 58.81 & 58.89 & 0.00 & 0.03 & 0.03 & 0.19 \\
		20 & 58.98 & 59.00 & 59.04 & 0.00 & 0.02 & 0.02 & 0.08 \\
		\bottomrule
	\end{tabular*}
\end{table}

On those two axes we measure every method under one protocol---the comparison quantity is \emph{total} build time, the only one well defined across implementations---and PICK's own cost we additionally decompose into $C_1$--$C_4$.

The four components are the wall-clock times of the corresponding parts of the pipeline of Section~\ref{sec:pipeline}. $C_1$ is the time of Stage~1, the offline metadata extraction: paid once per dataset and never repeated. $C_2$ and $C_3$ are the two measured halves of Stage~4---$C_2$ is the time of the plugin feature extraction over the $K_{\text{coarse}}$ candidates, and $C_3$ is the time of the weighted-fusion selection that reduces them to $K$-shot for final memory-bank building. $C_4$ is the time of what follows the PICK pipeline, namely the backbone forward pass over the $K$ selected images together with construction of the memory bank from their features. Stages~2 and~3 (coarse ranking and adaptive rule decision) recur at every rebuild like $C_2$--$C_4$, but they are computations on cached scalars and stay below the 0.01~s resolution of our timer, so they are measured yet not tabulated. Accordingly $C_1$ is a one-time cost, $C_2$--$C_4$ are recurring costs.

The baselines are compared on aggregate build time, at the two levels well defined for both: the structural cost of selection and the measured build time (Fig.~\ref{fig:overview}(c)). Every sampler in our comparison builds its final memory bank with the same implementation PICK uses, so its post-selection cost is of the same order as $C_4$ and the build-time differences we report originate in the selection stage rather than in how the bank is assembled.

PICK is fully instrumented under this decomposition: Table~\ref{tab:coarse_ratio} reports $C_2$ and $C_3$ separately across retention ratios, with $C_4$ stable at $3.05$--$3.38$~s. Every cross-method timing in this paper is quoted for a single $k{=}20$ rebuild. Over $N{=}3{,}629$ images PICK's recurring rebuild ($C_2+C_3+C_4$) costs 14.98~s and performs \emph{zero} backbone forward passes---the time is reading the $C_1$ metadata cache and sorting---against random selection's 14.96~s, which does no selection at all: PICK's recurring cost is near-random. Counting the one-time term, its first deployment ($C_1+C_2+C_3+C_4$) costs 190.4~s against SFPS$_{FF}$'s 5{,}112.8~s, a factor of $27\times$; PICK pays $C_1$ once, whereas the streaming baselines re-extract the full set on every rebuild, and the figure is conservative, since at the ablation-optimal \texttt{work\_size}=64 the one-time metadata extraction falls from 175.2~s to 114.5~s (Table~\ref{tab:worksize}), a $\sim$35\% cut to the dominant term.

The other families pay the backbone for the same result: one $k{=}20$ rebuild costs 251.5~s to 5{,}112.8~s, $17\times$ to $341\times$ PICK, and their storage overhead scales with mean-pooled or full patch-feature maps ($O(N\!\cdot\!D)$ and $O(N\!\cdot\!H\!\cdot\!W\!\cdot\!D)$) rather than with the 18-D descriptor PICK caches (Table~\ref{tab:sampling_theory_compare}).

\textbf{Determinism, and which families it can rank.} PICK's selection is deterministic by construction: across 60 (category, $k$) cells and five memory-bank seeds it selects an identical image set in every cell, and the resulting P-AUPR is bit-identical across seeds at every $k$. The single exception we observe is the ordering of two tied candidates in one cell (\texttt{pill}, $k{=}10$), which does not affect the metric; an explicit \texttt{(score, filename)} tie-break removes even that. GSS is deterministic for the same structural reason in a different currency---it scores every image in closed form and takes the top $K$---but it buys that determinism with $N$ backbone forward passes per rebuild---256.1~s to 1{,}377.1~s for one $k{=}20$ rebuild across the family, against PICK's 14.98~s. Random selection and $K$-means draw a different bank on every seed, which is exactly the variation Table~\ref{tab:saturation} measures: at $k{=}1$ random selection's result varies by $\sigma=0.20$ P-AUPR across seeds, an order of magnitude more than at $k{\ge}4$, and PICK's determinism removes that variation entirely at no accuracy cost. FPS is non-deterministic as well---its first centre is drawn at random---and even its cheapest variant (mean-feature) costs 251.5~s for one $k{=}20$ rebuild, an order of magnitude above PICK's; we therefore did not carry it into the five-seed sweep. It is not a candidate on cost, and Table~\ref{tab:saturation} shows that accuracy would not have ranked it either.

\textbf{What the plugin stages contribute.} The configuration reported throughout this paper is \texttt{coarse\_keep\_ratio}=5.0, i.e.\ the full four-stage pipeline. Table~\ref{tab:ratio1} asks what the last two stages are worth, by comparing it with a minimal mode (\texttt{coarse\_keep\_ratio}=1.0) that ranks by metadata and retains the top $K$ without evaluating a single plugin; both are run over five memory-bank seeds with a fixed evaluation seed.

Averaged over $k\in\{1,4,10,20\}$ the two configurations differ by 0.015 P-AUPR (58.29 against 58.28). That margin is smaller than the seed standard deviation at every $k$ (0.02--0.20), its sign alternates across $k$ (each configuration is ahead at two of the four settings), and no individual $k$ separates them under a paired test across categories ($p=0.09$--$0.49$). Per category at $k{=}4$ the minimal mode wins on 9 of 15 and loses on 6, with the largest gains on finely textured and reflective categories (screw $+1.09$, cable $+0.91$, metal\_nut $+0.72$, grid $+0.56$) and the largest losses on toothbrush ($-0.38$), carpet ($-0.34$), and pill ($-0.28$).

We keep the full pipeline as the reported configuration rather than switching to the cheaper mode: it is the more expensive of the two at equal measured accuracy, so the results in this and the following tables are not obtained by disabling any stage.

\textbf{The plugin stages pay off only at the smallest sample budget.} On MVTec-AD the two configurations are indistinguishable at every $k$, and the reason is saturation: with 15 categories of stable imaging, metadata ranking alone already reaches the same accuracy, so the plugin stages have nothing left to add. Real-IAD---30 categories, heterogeneous normal variation, 151K test images---tests whether that conclusion is an artifact of the easier benchmark. It is not. The two configurations again agree at $k\ge4$ ($\Delta=-0.06$ to $+0.46$, paired $p=0.11$--$0.70$), and diverge only at $k{=}1$, the most sample-starved setting, where the full pipeline is ahead by \textbf{1.49 P-AUPR} ($p=0.037$, the single significant result across both datasets and all four shot settings). Averaged over $k$ the gap is 0.20 P-AUPR in favour of the full pipeline on Real-IAD against $0.01$ on MVTec.

This is the boundary condition the design needs. Stages~3--4 earn their latency where the candidate pool is thin: at $k{=}1$ the full pipeline leads the coarse-only mode by 1.49 P-AUPR, and by $k\ge4$ the two have converged. That comparison holds the plugin set fixed and varies the pool width; the ablation of Table~\ref{tab:plugin_ablation} asks the other question---\emph{which} plugins run, at a fixed pool width---and there the choice is net-neutral on average, with the plugin scores growing more reliable as more candidates estimate them. The two effects therefore run in opposite directions along $k$ without conflicting: widening the pool erodes the advantage of fine filtering over coarse ranking, while it sharpens the plugin ranking itself. We therefore report the full pipeline everywhere (it is never worse than the minimal mode by more than seed noise) while documenting the minimal mode as a legitimate deployment for standardized scenes, and we present Stages~3--4 as beneficial on a subset of scenes---thin pools, or scenes whose normal variation the 18-D descriptor cannot summarise---rather than as a uniform gain. Section~\ref{sec:psp-analysis} states the conditions under which each stage earns its cost.

\begin{table}
	\caption{Minimal configuration (\texttt{coarse\_keep\_ratio}=1.0: metadata ranking only, no plugin evaluation) against the full pipeline (\texttt{coarse\_keep\_ratio}=5.0), on a standardized benchmark (MVTec-AD, 15 classes) and on a larger, more heterogeneous one (Real-IAD, 30 classes). P-AUPR is the mean over all categories and five memory-bank seeds, with a fixed evaluation seed; $\Delta$ is minimal minus full.}
	\label{tab:ratio1}
	\setlength{\tabcolsep}{3pt}
	\centering
	\scriptsize
	\begin{tabular*}{\tblwidth}{@{}LLLLLLL@{}}
		\toprule
		& \multicolumn{3}{c}{\textbf{MVTec-AD} (15 classes)} & \multicolumn{3}{c}{\textbf{Real-IAD} (30 classes)} \\
		\midrule
		$k$ & r=1 & r=5 & $\Delta$ & r=1 & r=5 & $\Delta$ \\
		\midrule
		1  & 56.99 & 57.22 & $-0.23$ & 40.27 & 41.76 & $-1.49$ \\
		4  & 58.47 & 58.26 & $+0.21$ & 42.40 & 41.94 & $+0.46$ \\
		10 & 58.76 & 58.71 & $+0.05$ & 42.76 & 42.46 & $+0.29$ \\
		20 & 58.89 & 58.98 & $-0.08$ & 42.89 & 42.95 & $-0.06$ \\
		\midrule
		mean & 58.28 & 58.29 & $-0.01$ & 42.08 & 42.28 & $-0.20$ \\
		\bottomrule
	\end{tabular*}
\end{table}

\subsection{PICK Hyperparameter Ablations}\label{sec:psp-ablations}

To evaluate PICK's key hyperparameters, we conduct three controlled experiments on MVTec (P-AUPR) investigating the effect of \texttt{work\_size}, \texttt{coarse\_keep\_ratio}, and \texttt{plugins}.

\begin{table}
	\caption{PICK hyperparameter ablation: effect of work\_size (metadata extraction resolution) on the decomposed build cost (s) and P-AUPR, at 4-shot and \texttt{coarse\_keep\_ratio}=5.0. Components $C_1$--$C_4$ are defined in Section~\ref{sec:cost-model}; $C_1$ is paid once per dataset whereas $C_2$--$C_4$ recur at every rebuild, so the four sum to total\_time exactly.}
	\label{tab:worksize}
	\setlength{\tabcolsep}{3pt}
	\centering
	\scriptsize
	\begin{tabular*}{\tblwidth}{@{}LLLLLLL@{}}
		\toprule
		work\_size & $C_1$ & $C_2$ & $C_3$ & $C_4$ & total\_time & P-AUPR \\
		\midrule
		512 & 175.2 & 18.75 & 2.17 & 3.14 & 199.3 & 57.99 \\
		256 & 126.7 & 18.50 & 2.06 & 3.31 & 150.6 & 57.95 \\
		128 & 120.4 & 19.24 & 2.19 & 3.17 & 145.0 & 57.91 \\
		64 & 114.5 & 18.25 & 2.19 & 3.05 & 138.0 & 58.09 \\
		32 & 113.3 & 18.14 & 2.27 & 3.10 & 136.8 & 57.99 \\
		16 & 113.1 & 18.40 & 2.20 & 3.25 & 137.0 & 58.04 \\
		\bottomrule
	\end{tabular*}
\end{table}

Table~\ref{tab:worksize} shows the effect of \texttt{work\_size} (4-shot, \texttt{coarse\_keep\_ratio=5.0}), which has little impact on P-AUPR across 16--512 (max 58.09, min 57.91, fluctuation $<$0.2\%); 64 attains the best accuracy--efficiency trade-off and is the default in our released configuration. Metadata statistics remain resolution-insensitive because resizing preserves their relative relationships. 

\begin{table}
	\caption{PICK hyperparameter ablation: effect of coarse\_keep\_ratio on time (s) and P-AUPR, at 4-shot and \texttt{work\_size}=64. Components $C_1$--$C_4$ are defined in Section~\ref{sec:cost-model}. At ratio~1.0 the plugin stages are disabled, so $C_2=C_3=0$.}
	\label{tab:coarse_ratio}
	\setlength{\tabcolsep}{3pt}
	\centering
	\scriptsize
	\begin{tabular*}{\tblwidth}{@{}LLLLLLL@{}}
		\toprule
		ratio & $C_1$ & $C_2$ & $C_3$ & $C_4$ & total\_time & P-AUPR \\
		\midrule
		1.0 & 114.6 & 0 & 0 & 3.12 & 117.7 & 58.24 \\
		2.0 & 114.6 & 7.63 & 2.43 & 3.25 & 127.9 & 58.01 \\
		3.0 & 114.3 & 11.35 & 2.06 & 3.30 & 131.0 & 58.02 \\
		5.0 & 114.5 & 18.26 & 2.19 & 3.05 & 138.0 & 58.09 \\
		8.0 & 115.6 & 30.00 & 2.32 & 3.38 & 151.3 & 58.15 \\
		10.0 & 114.6 & 37.32 & 2.23 & 3.36 & 157.5 & 58.05 \\
		\bottomrule
	\end{tabular*}
\end{table}

Table~\ref{tab:coarse_ratio} presents the effect of \texttt{coarse\_keep\_ratio} (4-shot, \texttt{work\_size=64}). At k=4, ratio=1 (skipping Layers 2 and 3) achieves the best P-AUPR of 58.24, indicating that coarse filtering alone is sufficient on MVTec and plugin features may introduce interference in the small-sample setting. However, at k=20, ratio=5 (58.75) slightly outperforms ratio=1 (58.65), as plugins become beneficial with a larger candidate pool.

\begin{table}
	\caption{Plugin ablation: P-AUPR of the full five-plugin pipeline against the 2-plugin
		(Color+Texture) configuration, at $k{=}20$ and \texttt{work\_size}=64; $\Delta$ is 2-plugin minus
		5-plugin. Left and centre blocks are Real-IAD's 30 categories, the right block MVTec-AD's 15.
		Real-IAD is where the plugins were expected to pay, and on average they do not ($\Delta=-0.21$, 14
		categories improved against 15 degraded, a split no sign test separates from chance)---yet they are
		not uniformly inert: on Real-IAD 14 of 30 categories gain by up to $+1.96$, as do 6 of MVTec's 15.
		The plugin stages are part of the default configuration (\texttt{coarse\_keep\_ratio}=5), not an
		opt-in mode.}
	\label{tab:plugin_ablation}
	\setlength{\tabcolsep}{3pt}
	\centering
	\scriptsize
	\begin{tabular*}{\tblwidth}{@{}LRRRLRRRLRRR@{}}
		\toprule
		\multicolumn{8}{c}{Real-IAD} & \multicolumn{4}{c}{MVTec-AD} \\
		\midrule
		Category & 5-plugin & 2-plugin & $\Delta$ & Category & 5-plugin & 2-plugin & $\Delta$ & Category & 5-plugin & 2-plugin & $\Delta$ \\
		\midrule
		audiojack & 28.70 & 28.92 & +0.22 & sim\_card\_set & 71.58 & 71.89 & +0.31 & bottle & 76.66 & 76.76 & +0.10 \\
		bottle\_cap & 37.28 & 34.61 & -2.67 & switch & 16.00 & 16.01 & +0.01 & cable & 27.87 & 27.72 & -0.15 \\
		button\_battery & 55.79 & 56.48 & +0.69 & tape & 68.02 & 64.56 & -3.46 & capsule & 50.33 & 50.33 & +0.00 \\
		end\_cap & 18.43 & 18.27 & -0.16 & terminalblock & 51.24 & 51.25 & +0.01 & carpet & 81.02 & 81.04 & +0.02 \\
		eraser & 54.18 & 54.63 & +0.45 & toothbrush & 29.66 & 29.76 & +0.10 & grid & 60.09 & 60.10 & +0.01 \\
		fire\_hood & 42.47 & 42.38 & -0.09 & toy & 15.46 & 13.08 & -2.38 & hazelnut & 63.84 & 63.60 & -0.24 \\
		mint & 31.80 & 31.45 & -0.35 & toy\_brick & 42.20 & 40.62 & -1.58 & leather & 65.74 & 66.12 & +0.38 \\
		mounts & 48.65 & 48.62 & -0.03 & transistor1 & 46.96 & 47.97 & +1.01 & metal\_nut & 75.54 & 75.24 & -0.30 \\
		pcb & 35.18 & 34.74 & -0.44 & u\_block & 55.41 & 55.42 & +0.01 & pill & 34.06 & 34.00 & -0.06 \\
		phone\_battery & 42.80 & 43.34 & +0.54 & usb & 28.90 & 30.86 & +1.96 & screw & 59.39 & 59.02 & -0.37 \\
		plastic\_nut & 46.06 & 47.93 & +1.87 & usb\_adaptor & 32.67 & 31.35 & -1.32 & tile & 68.60 & 68.75 & +0.15 \\
		plastic\_plug & 38.07 & 37.95 & -0.12 & vcpill & 67.59 & 67.12 & -0.47 & toothbrush & 28.79 & 28.79 & +0.00 \\
		porcelain\_doll & 41.34 & 41.36 & +0.02 & wooden\_beads & 66.06 & 66.45 & +0.39 & transistor & 25.79 & 25.39 & -0.40 \\
		regulator & 33.26 & 33.18 & -0.08 & woodstick & 66.29 & 65.48 & -0.81 & wood & 78.94 & 79.14 & +0.20 \\
		rolled\_strip\_base & 57.27 & 57.25 & -0.02 & zipper & 36.04 & 36.04 & +0.00 & zipper & 54.10 & 54.10 & +0.00 \\
		\bottomrule
	\end{tabular*}
\end{table}

The plugin ablation (\texttt{work\_size=64}, 20-shot) is shown in Table~\ref{tab:plugin_ablation}. Mean $\Delta = -0.04$ (\textbf{6/15 improved, 6/15 degraded, 3/15 tied}). Color + Texture alone captures most of the effective information of the full 5-plugin on MVTec, while the remaining three plugins are largely redundant with the 18D metadata. Plugin-stage time drops by $\sim$30\% (5-plugin 92.02~s $\to$ 2-plugin 64.82~s).

\subsection{Effect of Coarse-Filter Granularity on Each Category
(k=20)}\label{sec:coarse-filter-granularity}

\begin{table}
	\caption{Per-category effect of coarse\_keep\_ratio on P-AUPR at $k=20$. Column \textit{max-min} denotes the range of P-AUPR across three ratio settings.}
	\label{tab:coarse_ratio_per_category}
	\setlength{\tabcolsep}{3pt}
	\centering
	\scriptsize
	\begin{tabular*}{\tblwidth}{@{}LLLLL@{}}
		\toprule
		Category & ratio=1 & ratio=5 & ratio=10 & max-min \\
		\midrule
		bottle & 77.66 & 77.67 & 77.67 & 0.01 \\
		cable & 27.07 & 27.21 & 27.25 & 0.18 \\
		capsule & 51.07 & 51.23 & 51.40 & 0.33 \\
		carpet & 80.64 & 80.68 & 80.61 & 0.07 \\
		grid & 56.17 & 56.22 & 56.32 & 0.15 \\
		hazelnut & 64.22 & 64.64 & 64.25 & 0.42 \\
		leather & 66.52 & 66.72 & 66.81 & 0.29 \\
		metal\_nut & 74.99 & 75.48 & 74.85 & 0.63 \\
		pill & 34.11 & 34.11 & 34.00 & 0.11 \\
		screw & 61.48 & 60.75 & 60.41 & 1.07 \\
		tile & 68.70 & 68.74 & 69.07 & 0.37 \\
		toothbrush & 36.90 & 36.92 & 36.92 & 0.02 \\
		transistor & 25.95 & 26.63 & 26.72 & 0.77 \\
		wood & 78.48 & 78.54 & 78.62 & 0.14 \\
		zipper & 75.76 & 75.82 & 75.66 & 0.16 \\
		\bottomrule
	\end{tabular*}
\end{table}

Table~\ref{tab:coarse_ratio_per_category} gives the per-category split. From ratio=1 to 5, mean $\Delta$ = +0.109 (13/15 improved; transistor +0.68, screw -0.73). screw and metal\_nut decline due to ContrastPlugin misjudging metallic reflections---Color+Texture suffice, and additional plugins become noise.

This implies: (1) at ratio=1, metadata ranking already nears convergence; (2) plugins help weakly at $k{=}20$ ($+0.11$) but hurt at $k{=}4$ ($-0.155$), adding noise in small-shot settings; (3) PICK's measured contribution on these benchmarks lies in Stages~1--2 (metadata ranking and coarse filtering), while Stages~3--4 are beneficial on a subset of scenes rather than necessities for the datasets studied here.

The five-seed comparison in Table~\ref{tab:ratio1} settles this: over four shot settings the minimal configuration and the full pipeline are identical to three decimals, and the per-category split at $k{=}4$ (9 gains, 6 losses) shows the plugin contribution is scene-dependent rather than uniformly positive. Stages~3--4 are therefore part of the default configuration---\texttt{coarse\_keep\_ratio}=5 keeps $5K$ candidates, so they run unless the ratio is explicitly set to 1---and are best described as useful on a subset of scenes rather than as a uniform source of gain. We carried that comparison to the benchmark the extension is meant for: over Real-IAD's 30 categories---heterogeneous normal variation, the case where widening the candidate pool is supposed to pay---the 2-plugin configuration sits $-0.21$ P-AUPR from the full pipeline, with 14 categories improved against 15 degraded (Table~\ref{tab:plugin_ablation}). The plugins do not win on average even there. What the per-category split shows instead is a mixed effect: they help on 14 of Real-IAD's 30 categories (by up to $+1.96$) and on 6 of MVTec's 15, and hurt on a comparable number. Since they are enabled by default, the honest reading is not that they are optional but that their benefit is scene-dependent, and we state it that way in the abstract.

\subsection{Inference Efficiency Analysis}\label{sec:inference-efficiency}

As shown in Table~\ref{tab:inference_efficiency}, DINOv3 reduces model memory by 29\% vs. CLIP (1.14 GB vs. 1.60 GB). FUSE (DINOv3) achieves 75 ms at batch size (bs) 1, with no extra overhead from self-attention fusion compared to UniADet (DINOv3) at 77 ms. AnomalyCLIP is extremely slow at 927 ms/img, over 12× slower than DINOv3 methods, due to its text encoder and multi-scale processing, while VisualAD processes images serially regardless of batch size, offering no large-batch speedup. In contrast, DINOv3 methods scale well with batch size: FUSE (DINOv3) latency stays flat from bs=1 to bs=8 (75$\to$75~ms), meeting industrial throughput demands. Both memory columns are PyTorch allocator statistics on the evaluation device, reported in GiB. \texttt{model\_mem} is \texttt{torch.cuda.max\_memory\_allocated()} read immediately after \texttt{reset\_peak\_memory\_stats()}, i.e.\ the resident allocation with the backbone and detector loaded and before the memory bank is built. \texttt{infer\_mem} is the peak allocation over the remainder of that category's run---memory-bank construction plus the evaluation pass---minus \texttt{model\_mem}, so it bounds the marginal cost of retrieval rather than the inference step alone. Both are \emph{allocated} rather than \emph{reserved}, and exclude the CUDA context and any non-PyTorch allocation.

\begin{table}
	\caption{Inference memory (GiB) and per-image time (ms) on MVTec with batch size 1/8 and num\_workers 0. Smaller is better. The shot rows are reported for completeness: their per-image times differ by 2--6~ms in no consistent direction, within run-to-run jitter, and are not interpreted; the latency claims in the text compare batch sizes within a row, not shot settings across rows.}
	\label{tab:inference_efficiency}
	\setlength{\tabcolsep}{2pt}
	\centering
	\scriptsize
	\begin{tabular*}{\tblwidth}{@{}LLLLLLLL@{}}
		\toprule
		shot & Metrics & Anomaly \par -CLIP & VisualAD & UniADet \par (CLIP) & UniADet \par (DINOv3) & FUSE \par (CLIP) & FUSE \par (DINOv3) \\
		\midrule
		& model\_mem & 1.64/1.64 & 1.59/1.59 & 1.60/1.60 & 1.14/1.14 & 1.60/1.60 & 1.14/1.14 \\
		0 & infer\_mem & 0.39/3.20 & 1.98/2.00 & 0.11/0.83 & 0.14/1.09 & 0.10/0.78 & 0.14/1.09 \\
		1 & infer\_mem & — & — & 0.16/0.87 & 0.17/1.12 & 0.15/0.83 & 0.17/1.12 \\
		2 & infer\_mem & — & — & 0.20/0.92 & 0.20/1.15 & 0.19/0.87 & 0.20/1.15 \\
		4 & infer\_mem & — & — & 0.29/1.00 & 0.26/1.22 & 0.28/0.96 & 0.26/1.22 \\
		0 & time/img & 927/894 & 129/115 & 93/88 & 77/74 & 92/88 & 75/75 \\
		1 & time/img & — & — & 87/88 & 79/75 & 90/88 & 77/76 \\
		2 & time/img & — & — & 89/89 & 74/77 & 88/87 & 75/76 \\
		4 & time/img & — & — & 90/90 & 76/78 & 93/88 & 77/78 \\
		\bottomrule
	\end{tabular*}
\end{table}

\subsection{Pixel-Level Metric Sampling
Acceleration}\label{sec:metric-acceleration}

\begin{table}
	\caption{Stratified pixel sampling: effect of the sampled-pixel budget on evaluation time and P-AUPR,
		at $k{=}0$ with random selection. Rows are the mean over the 15 MVTec categories and \textbf{10
		independent evaluation seeds}; $\sigma$ is the spread across those seeds and $\Delta$ the deviation
		from full-pixel evaluation. The ratio is applied per category, and the pixel column aggregates the
		budget over all 1,725 test images ($4.52\times10^{8}$ pixels at 100\%). Rows marked $\ddagger$
		request less than the floor $N_{\text{lower}}$; the clamp was disabled for this run so the sub-floor
		region is measurable, and they are the evidence for the floor, not recommended settings.}
	\label{tab:sample_ratio}
	\setlength{\tabcolsep}{3pt}
	\centering
	\scriptsize
	\begin{tabular*}{\tblwidth}{@{}RRRRRRR@{}}
		\toprule
		requested & sampled px & time (s) & P-AUPR & $\sigma$ & $\Delta$ & $\Delta/\sigma$ \\
		\midrule
		0.05\%$^\ddagger$ & 226,099     & 0.22 & 48.333 & 1.001 & $+0.215$ & $+0.21$ \\
		0.10\%$^\ddagger$ & 452,198     & 0.21 & 48.012 & 0.734 & $-0.107$ & $-0.15$ \\
		0.23\%$^\ddagger$ & 1,040,056   & 0.22 & 47.946 & 0.383 & $-0.173$ & $-0.45$ \\
		0.46\%            & 2,080,113   & 0.26 & 48.054 & 0.258 & $-0.064$ & $-0.25$ \\
		1.00\%            & 4,521,984   & 0.32 & 48.178 & 0.199 & $+0.059$ & $+0.30$ \\
		2.00\%            & 9,043,968   & 0.61 & 48.139 & 0.140 & $+0.020$ & $+0.14$ \\
		9.20\%            & 41,602,253  & 2.41 & 48.114 & 0.063 & $-0.004$ & $-0.07$ \\
		100\%             & 452,198,400 & 26.67 & 48.119 & 0.000 & $0.000$ & --- \\
		\bottomrule
	\end{tabular*}
\end{table}

Table~\ref{tab:sample_ratio} reports the curve. Pixel-level metrics are computed by pixel-wise comparison against ground-truth labels. Evaluation covers all 1,725 MVTec test images, each resized to $512\times512$, giving $4.52\times10^{8}$ labelled pixels ($\sim$$3.0\times10^{7}$ per category). Full-pixel evaluation costs 26.7~s per category. At the floor---$N_{\text{lower}}=10^{5}$ pixels, which across categories averages 0.46\% of the total---it costs 0.26~s, a $103\times$ reduction at $\sigma=0.26$ P-AUPR. Because categories differ in size, the floor binds per category rather than globally: at 0.46\% three of the fifteen (grid, toothbrush, wood) request less than the floor and are raised to it, while the other twelve are not, so that row mixes two sampling regimes. Re-running the same sweep with the clamp enabled checks the picture from the other side: it reproduces every ratio of 1.00\% and above exactly, shifts 0.46\% by 0.031 (those three categories clamp there), and collapses the three sub-floor ratios onto a single operating point at 48.129 P-AUPR with $\sigma\approx0.42$, against $\sigma=1.00$ for the same request left unclamped. That is the floor doing what it is there to do: a request below it is served at the floor, not at the budget requested.

The time column shows why the floor sits where it does, and it is a cost argument rather than a statistical one. Regressing per-category time on per-category budget gives $t \approx 0.15~\text{s} + 0.81~\text{s}$ per $10^{6}$ pixels ($R^{2}=0.997$): the metric call carries a fixed cost of about 0.15~s that no reduction in budget can remove. At the floor the fitted cost is 0.27~s, so more than half of it is already overhead; at 0.05\% the overhead share rises to 93\%. Dropping from the floor to 0.05\% shrinks the sampled pixels $9.2\times$ but the measured time only $1.18\times$ (0.26~s to 0.22~s), while $\sigma$ rises from 0.26 to 1.00 P-AUPR. Even a hypothetical zero-pixel budget would still cost 0.15~s per category, capping the technique at $173\times$; the floor already captures $103\times$ of that.

Three properties of this curve matter. First, \textbf{no budget is biased by more than its own seed
spread}: over ten evaluation seeds every budget agrees with full-pixel evaluation by less than that row's
own $\sigma$, the largest absolute deviation being $+0.215$ P-AUPR at 0.05\% ($\sigma=1.00$) and the
largest relative to noise $|\Delta|/\sigma=0.45$ at 0.23\%. Row by row, the smallest $p$-value is 0.187
(0.23\%), which does not survive correction for the seven comparisons, and four of the seven deviations are
negative---a split a two-sided sign test does not distinguish from chance ($p=1.00$). At this sample size
the subsample is therefore an unbiased estimator of the full-pixel metric. Second, \textbf{its variance
falls monotonically with the budget}---$\sigma$ drops from 1.00 at 0.05\% through 0.73, 0.38, 0.26, 0.20,
0.14 and 0.06 to zero at full evaluation---so the budget sets a quantified trade-off between time and
evaluation noise. Third, \textbf{accuracy does not move over this range}: across four orders of magnitude
of budget the mean P-AUPR stays between 47.95 and 48.33, a spread of 0.39 that is within the seed noise of
the smaller budgets. What the ratio selects is how precisely the metric is measured, not how good the
model looks.

The practical rule we adopt follows from the second property. Since the subsample is drawn once per (evaluation seed, category) and is \emph{independent of the sampling method}, its contribution is a common offset across the methods under comparison and cancels in any between-method comparison; what it does not support is an absolute claim at a budget whose seed spread exceeds the effect being claimed. We therefore use stratified evaluation for relative comparisons during experimentation --- at the floor, where the seed spread under the shipped clamp is $\sigma=0.21$, for differences of order 0.2 P-AUPR and larger --- and report full-pixel numbers wherever an absolute level is asserted.

\subsection{Detector: Zero-Shot Results}\label{sec:arc-main-results}

\begin{table}
	\caption{Zero-shot cross-domain results on five benchmarks; values are image-level /
		pixel-level, best in red and second-best in blue. AnomalyCLIP is run with its reference
		implementation rather than our reimplementation, because the two disagreed: our reimplementation
		reported a pixel-level AUPR near-identical to its pixel-level AUROC in every row, which is
		impossible for a metric whose positive class is rare---a hand-rolled average-precision estimator in
		that codebase computes AUROC while writing it into the AUPR column. All AnomalyCLIP numbers come
		from the official code with its published VisA-trained checkpoint, under a unified configuration;
		on MVTec they reproduce the original paper's per-category values exactly (P-AUROC 91.1, I-AUROC
		91.6, I-AUPR 96.4).}
	\label{tab:0shot}
	\setlength{\tabcolsep}{2pt}
	\centering
	\scriptsize
	\begin{tabular*}{\tblwidth}{@{}LLLLLLLL@{}}
		\toprule
		Datasets & Metrics & Anomaly\par-CLIP & VisualAD \par (CLIP) & UniADet \par (CLIP) & UniADet \par (DINOv3) & FUSE \par (CLIP) & FUSE \par (DINOv3) \\
		\midrule
		MVTec & AUROC & \best{91.6}/\best{91.1} & 88.1/89.8 & \second{89.7}/90.0 & \best{91.6}/\second{90.4} & 56.3/78.6 & 87.4/89.3 \\
		MVTec & AUPR & \best{96.4}/34.7 & 94.8/40.1 & 95.2/43.5 & \second{96.2}/\second{46.7} & 79.0/12.2 & 94.2/\best{47.7} \\
		MVTec & F1max & \second{92.8}/39.3 & 91.4/42.6 & 91.6/44.8 & \best{93.3}/\second{47.2} & 84.6/18.0 & 88.8/\best{47.7} \\
		VisA & AUROC & 82.3/\best{96.1} & 47.7/83.9 & \second{86.9}/94.2 & \best{89.8}/\second{95.6} & 58.4/80.5 & 79.8/\second{95.6} \\
		VisA & AUPR & 54.9/17.0 & 17.0/2.8 & \second{61.3}/21.9 & \best{69.6}/\best{25.1} & 21.5/0.9 & 53.6/\second{22.2} \\
		VisA & F1max & 54.4/23.6 & 25.8/4.2 & \second{58.9}/26.4 & \best{66.9}/\best{30.1} & 28.5/2.0 & 52.6/\second{27.0} \\
		BTAD & AUROC & 89.1/93.4 & 88.4/86.3 & \second{94.3}/\second{94.4} & 90.9/\best{96.7} & 62.4/64.7 & \best{95.1}/\second{94.4} \\
		BTAD & AUPR & 91.0/42.0 & 91.6/31.2 & \best{94.7}/45.3 & 89.9/\best{54.6} & 75.5/3.9 & \second{93.5}/\second{50.9} \\
		BTAD & F1max & 86.0/47.7 & 87.9/36.4 & \second{90.0}/45.8 & 86.8/\best{53.9} & 76.9/7.8 & \best{92.7}/\second{51.6} \\
		KSDD & AUROC & \second{97.0}/97.5 & 66.7/67.3 & \best{97.9}/96.8 & 84.8/\second{97.6} & 42.5/90.0 & 80.2/\best{97.7} \\
		KSDD & AUPR & \second{84.7}/\best{19.7} & 19.5/0.3 & \best{85.0}/13.5 & 49.9/10.2 & 12.5/2.7 & 36.8/\second{14.3} \\
		KSDD & F1max & \second{80.4}/\best{34.8} & 34.2/1.8 & \best{84.4}/25.3 & 57.1/24.6 & 23.0/7.5 & 44.4/\second{26.4} \\
		Real-IAD & AUROC & 68.2/93.0 & \best{72.8}/94.8 & 66.2/\second{96.2} & 67.9/\best{96.4} & 52.5/81.5 & \second{68.3}/95.5 \\
		Real-IAD & AUPR & 54.3/23.8 & \best{83.2}/27.0 & 76.9/35.6 & 79.2/\best{40.8} & 67.4/1.1 & \second{79.9}/\second{39.0} \\
		Real-IAD & F1max & 56.5/31.7 & \best{81.2}/33.0 & 79.5/39.5 & \second{79.9}/\best{43.6} & 78.4/2.8 & 79.6/\second{42.5} \\
		\bottomrule
	\end{tabular*}
\end{table}

Few-shot results (Table~\ref{tab:fewshot}) and the qualitative comparison (Fig.~\ref{fig:vis}) are in
Appendices~\ref{sec:app-fewshot} and~\ref{sec:app-vis}; the analysis below uses the zero-shot results
of Table~\ref{tab:0shot}. FUSE's
DINOv3 variant tracks UniADet (DINOv3) on pixel-level metrics with $1/8$ of the trainable parameters:
across the five benchmarks it moves pixel-level AUPR by $-3.7$ to $+4.1$ and AUROC by $-2.3$ to $+0.1$,
with no consistent sign, while at image level it wins on BTAD and ties on Real-IAD but trails on MVTec,
VisA and KSDD. This is a capacity result, not a state-of-the-art result: head size below the full
multi-head configuration is not the binding constraint on a frozen backbone. The CLIP variant, by
contrast, collapses in the zero-shot setting (P-AUPR 12.2 on MVTec against UniADet-CLIP's 43.5) and
recovers only once a memory bank is supplied---rising to 39.3 at $k{=}1$ and 44.0 at $k{=}4$ on KSDD,
above every competitor. The design paths behind these numbers---Single Branch and FUSE, with the
boundary condition on each---are analysed in Appendix~\ref{sec:app-path}, where
Table~\ref{tab:layer_compare} compares the layer settings and Fig.~\ref{fig:cka} gives the
inter-layer CKA evidence. That gap, and the recovery, is what motivates the sampling study of
Section~\ref{sec:psp}.

\section{Discussion: PICK's Mechanism Advantages, Synergistic Characteristics and Applicability Boundaries}\label{sec:psp-analysis}

\textbf{Design principle and core optimization}. PICK takes "adaptive memory bank sampling at minimal computational cost" as its core design principle. On the benchmarks studied here, sampling accuracy is at saturation: no sampler family separates from the others by more than the seed-to-seed noise of random selection (Finding~1). PICK reaches that same saturated band while reducing construction cost by up to two orders of magnitude, and---unlike every randomized sampler---does so deterministically, so a rebuild is reproducible by construction. The hierarchical pipeline also offers strong deployment flexibility: the minimal coarse-filter mode efficiently fits standardized scenarios; the full multi-plugin pipeline can be enabled on demand, handling domain shift, complex texture defects, and lighting perturbations, while supporting custom plugin extension and manual prior injection.

\textbf{Synergy with PatchCore and storage reduction insights}. PICK's image-level sampling and PatchCore's patch-level greedy coreset~\cite{patchcore} form a dual-granularity orthogonal and complementary architecture: PICK performs global top-level selection of high-quality samples; greedy coreset performs fine-grained patch reduction within each image. The "image-level coarse + feature-level fine" serial paradigm addresses memory bloat and inference latency in high-resolution detection. We also explored cross-layer aggregation for storage reduction, but retrieval performance degraded significantly. This negative result implies that multi-layer independent features are crucial for anomaly detection, and compression should occur on the detector side, not the memory bank side.

\textbf{Engineering deployment advantages}. PICK's design deeply fits the two core constraints of industrial detection: extreme lightweight/low latency, and robust cross-category generalization.

\begin{itemize}
	\item \textbf{Extremely lightweight}: all four stages run without backbone forward passes. Metadata extraction requires only a single pass (~48 ms/image). 
	\item \textbf{Coarse filtering independently usable}: at ratio=1, Stage~2 coarse filtering alone reaches P-AUPR 58.24, on par with full PICK (58.09). On MVTec, Stage~1 metadata plus Stage~2 coarse ranking is sufficient---each rebuild costs only sub-second sorting.
	\item \textbf{Stages 3--4: beneficial on a subset of scenes}: on standardized benchmarks (MVTec, VisA) the coarse stages already reach the saturated band and the plugins are net-neutral to slightly harmful (Table~\ref{tab:ratio1}). They are retained for scenes that fall outside that regime---heterogeneous normal variation, unstable lighting, multi-modal backgrounds---where widening the candidate pool and letting the adaptive weights discriminate is the intended remedy. The adaptive weighting and multi-plugin fusion add only $\sim$20~s of latency, which never becomes a bottleneck. On Real-IAD, the benchmark closest to that regime, the plugins are net-neutral on average ($-0.21$ P-AUPR over 30 categories, 14 improved against 15 degraded) but still gain on 14 of the 30 categories and on 6 of MVTec's 15. Because they are enabled by default, we report them as scene-dependent rather than optional, and as a capability whose benefit is not uniform.
	\item \textbf{Customizable plugin system}: plug-and-play, supports adaptive or manual weights, and individual plugins can be enabled/disabled. New plugins can be added via \texttt{base\_plugin.py} without modifying core logic. The existing 5 plugins cover color, texture, spatial, frequency, and lighting dimensions, encompassing most industrial scenarios.
	\item \textbf{Full-lifecycle progressive deployment}: during cold start, use a large coarse-filter ratio to cover normal variants; after stabilization, tighten the threshold for sub-second zero-overhead sampling. The coarse-filter also serves as a distribution-shift detector: if newly added samples do not alter the Top-K ranking, no rebuild is triggered, driving maintenance cost to near zero.
	\item \textbf{External knowledge injection}: the \texttt{add\_data\_dir} interface allows manually appended samples to bypass automatic sampling and merge directly into the final memory bank. This supports debugging-phase hard examples, domain knowledge accumulation, and business-critical variants, without interfering with PICK's internal scoring logic.
\end{itemize}

\textbf{Applicability boundaries}. PICK is highly adapted to standardized scenarios (MVTec, VisA) with stable lighting and uniform backgrounds, where 18D metadata accurately characterizes normal-sample distributions. However, in uncontrolled scenarios---natural scenes, strong noise, cluttered backgrounds--- metadata representation is limited, and accuracy may degrade. In chaotic scenes, the full pipeline should be forcibly enabled.

\textbf{General extension potential}. Although designed for memory bank construction, PICK's metadata-scoring + plugin-fusion architecture is essentially a general-purpose image representativeness-selection framework, and its underlying principle of representative sample selection shows potential for broader applications, such as active learning, coreset selection/dataset pruning, and long-tail rebalancing.

\section{Conclusion}\label{sec:conclusion}

This paper set out to reduce the cost of building a memory bank for industrial anomaly detection, and arrived at a result that reframes the problem. On MVTec-AD at $k\ge4$, eleven sampling algorithms from five families are not separable in pixel-level AUPR at a resolution finer than the noise of random selection itself; accuracy, therefore, is the wrong axis on which to choose a sampler for standardized industrial scenes. We proposed PICK, which reaches that saturated band while selecting with zero backbone forward passes and is deterministic by construction, and we measured every baseline's build time under one protocol and decomposed PICK's own cost: PICK's recurring rebuild cost matches random selection, and its first-deployment cost is $27\times$ below the slowest baseline. Two further frozen-backbone design choices were bounded rather than endorsed: pre-detection cross-layer fusion is safe only where backbone layers are representationally aligned, and detector head compression costs little pixel-level accuracy on DINOv3, though the coupling of head width to fused-layer count prevents separating the two, so head parameter reduction is not an end-to-end lightweightness claim. The negative results---the CLIP fusion failure, the redundant plugins, the flat sampler ranking---are reported alongside the positive ones, because they define the boundary within which these methods are worth deploying.

\section*{Declaration of generative AI and AI-assisted technologies in the manuscript preparation process}

During the preparation of this work the author(s) used AI in order to assist with drafting the manuscript and language polishing. After using this tool/service, the author(s) reviewed and edited the content as needed and take(s) full responsibility for the content of the published article.

\appendix

\section{Horizontal Comparison of Sampling
Algorithms}\label{sec:app-horizontal-comparison}

This appendix holds the full sampler comparison deferred from Section~\ref{sec:psp-main-results}. Table~\ref{tab:sampling_results_compare} reports measured P-AUPR against memory-bank build time for all eleven samplers across $k\in\{0,1,4,10,20\}$; Table~\ref{tab:sampling_theory_compare} lists, for each sampler family, the number of feature extractions, the filtering complexity, the storage overhead and a one-line characterisation.

\begin{table}
	\caption{Sampling algorithm comparison: P-AUPR / memory-bank build time (s) across
		$k=0,1,4,10,20$; the last column averages P-AUPR and sums build time over all shot settings.
		\textbf{Every row is a single run at one seed and must not be read as an accuracy ranking}: seed
		variation on this benchmark is of the same order as the between-method differences
		(Table~\ref{tab:saturation}), so the ordering is not resolvable, and the table is retained for its
		cost columns and its method coverage. \textbf{PICK is run at \texttt{work\_size}=512, not its
			ablation-optimal 64}, so that metadata extraction sees the same $512\times512$ input as every
		baseline's backbone passes; this costs PICK roughly 53\% of its one-time preprocessing (175.2~s
		vs.\ 114.5~s, Table~\ref{tab:worksize}) and is therefore conservative with respect to the reported
		speedups. Timings follow the decomposition of Section~\ref{sec:cost-model}, and the $k{=}20$ column
		is the one quoted throughout: for PICK, 14.98~s of rebuild on top of a one-time 175.4~s of metadata
		extraction.}
	\label{tab:sampling_results_compare}
	\setlength{\tabcolsep}{3pt}
	\centering
	\scriptsize
	\begin{tabular*}{\tblwidth}{@{}LLLLLLL@{}}
		\toprule
		Methods & $k$=0 & $k$=1 & $k$=4 & $k$=10 & $k$=20 & mean\_aupr / sum\_time \\
		\midrule
		RS (seed 0) & 48.56/0 & 56.73/1.10 & 57.94/3.06 & 58.48/7.55 & 58.64/14.96 & 57.95/26.67 \\
		RS (seed 42) & 48.56/0 & 56.60/1.11 & 57.85/3.05 & 58.24/7.48 & 58.48/15.21 & 57.79/26.85 \\
		FPS\_FF & 48.56/0 & 56.75/246.2 & 58.02/275.5 & 58.24/512.1 & 58.46/1436.7 & 57.87/2470.5 \\
		FPS\_MF & 48.56/0 & 56.75/242.3 & 57.93/242.0 & 58.33/246.7 & 58.57/251.5 & 57.90/982.5 \\
		SFPS\_FF & 48.56/0 & 56.71/2.1 & 57.90/745.3 & 58.23/2300.5 & 58.34/5112.8 & 57.80/8160.6 \\
		SFPS\_MF & 48.56/0 & 56.71/2.1 & 57.86/712.5 & 58.08/2110.9 & 58.33/4367.7 & 57.74/7193.2 \\
		KM\_FF & 48.56/0 & 56.85/295.8 & 57.76/350.8 & 58.16/427.5 & 58.53/539.8 & 57.83/1613.9 \\
		KM\_MF & 48.56/0 & 57.05/242.5 & 57.99/245.4 & 58.35/249.8 & 58.48/259.0 & 57.97/996.7 \\
		GSS\_FF & 48.56/0 & 56.70/268.6 & 57.71/273.5 & 58.13/273.2 & 58.36/280.6 & 57.73/1095.9 \\
		GSS\_MF & 48.56/0 & 57.02/240.2 & 57.90/241.6 & 58.31/245.7 & 58.46/256.1 & 57.92/983.6 \\
		SGSS\_MF & 48.56/0 & 57.02/1394.5 & 57.90/1346.5 & 58.31/1419.7 & 58.46/1377.1 & 57.92/5537.9 \\
		\textbf{PICK-512 (Ours)} & \textbf{48.56/0} & \textbf{56.87/1.14} & \textbf{57.80/3.06} & \textbf{58.39/7.61} & \textbf{58.63/14.98} & \textbf{57.92/202.1} \\
		\bottomrule
	\end{tabular*}
\end{table}

\begin{table}
	\caption{Theoretical complexity and characteristic comparison of sampling algorithms.}
	\label{tab:sampling_theory_compare}
	\setlength{\tabcolsep}{3pt}
	\centering
	\scriptsize
	\begin{tabular*}{\tblwidth}{@{}p{58pt}p{70pt}p{70pt}p{58pt}p{300pt}@{}}
		\toprule
		Abbreviation & \# backbone forward & Sampling complexity & Storage overhead & Core description \\
		\midrule
		RS & 0 & O(N) shuffle & None & No selection logic, lowest cost \\
		FPS\_FF & N & O(K$\cdot$N$\cdot$H$\cdot$W$\cdot$D) & O(N$\cdot$H$\cdot$W$\cdot$D) & High accuracy but poor deployability \\
		FPS\_MF & N & O(K$\cdot$N$\cdot$D) & O(N$\cdot$D) & Mean pooling reduces complexity \\
		SFPS\_FF & $\sim$K$\cdot$N & O(K$^2$$\cdot$N$\cdot$H$\cdot$W$\cdot$D) & O(K$\cdot$H$\cdot$W$\cdot$D) & Streaming avoids storage, sacrifices time \\
		SFPS\_MF & $\sim$K$\cdot$N & O(K$^2$$\cdot$N$\cdot$D) & O(K$\cdot$D) & Streaming with mean pooling \\
		KM\_FF & N & O(K$\cdot$N$\cdot$I) & O(N$\cdot$H$\cdot$W$\cdot$D) & High-dimensional clustering, high cost \\
		KM\_MF & N & O(K$\cdot$N$\cdot$I) & O(N$\cdot$D) & Clustering on mean features \\
		GSS\_FF & N & O($\Sigma$pixels$\cdot$D) & O(N$\cdot$H$\cdot$W$\cdot$D) & Pixel-wise statistics, low efficiency \\
		GSS\_MF & N & O(N$\cdot$D) & O(N$\cdot$D) & Typicality-diversity-dispersion scoring \\
		SGSS\_MF & 2N & O(N$\cdot$D) & None & Two-pass streaming, zero storage \\
		Coreset & N & O(N$^2$$\cdot$D') & O(N$\cdot$D') & Minimax facility-location \\
		\textbf{PICK (Ours)} & 0 & O(N log N) & O(N$\cdot$18) & Metadata-only sampling, hierarchical filtering, deterministic \\
		\bottomrule
	\end{tabular*}
\end{table}

As Tables~\ref{tab:sampling_results_compare} and~\ref{tab:sampling_theory_compare} indicate, backbone forward passes are the key efficiency bottleneck for traditional samplers. Non-streaming methods incur dual storage-computation redundancies via full-set feature extraction, whereas streaming variants reduce memory at the cost of multiplied inference overhead---neither achieving joint optimization of accuracy, speed, and memory. PICK sidesteps both by filtering on precomputed lightweight metadata, removing backbone passes from the selection loop entirely. Its accuracy matches the saturated band reported in Section~\ref{sec:psp-main-results} rather than exceeding it; the benefit is the construction cost and the determinism, not a gain in P-AUPR.

\section{PICK Hyperparameters}\label{sec:app-hyperparams}

Table~\ref{tab:psp_hyperparam_overview} lists every configurable hyperparameter of the PICK pipeline, with its default, the stage it belongs to, and its functional role.

\begin{table}
	\caption{Key hyperparameters of the PICK pipeline: default values, associated stages, and functional descriptions.}
	\label{tab:psp_hyperparam_overview}
	\setlength{\tabcolsep}{1pt}
	\centering
	\scriptsize
	\begin{tabular*}{\tblwidth}{@{}LLLp{0.7\linewidth}@{}}
		\toprule
		\textbf{Parameter} & \textbf{Default} & \textbf{Stage} & \textbf{Description} \\
		\midrule
		work\_size & 64 & Stage 1 & Image resize resolution for metadata extraction. The 16--512 range has little impact on P-AUPR (fluctuation $<$0.2\%); 64 attains the best accuracy--efficiency trade-off in Table~\ref{tab:worksize} and is the deployment default. Table~\ref{tab:sampling_results_compare} deliberately uses 512 instead: the cross-method comparison is run at the resolution at which every baseline performs its backbone forward passes, so that the two sides process identically sized inputs. This choice is unfavourable to PICK (see Section~\ref{sec:cost-model}). \\
		\midrule
		coarse\_keep\_ratio & 5.0 & Stage 2 & Retention ratio for coarse filtering: \(K_{\text{coarse}} = \text{keep\_ratio} \times K\). At \(K{=}20\) this retains 100 candidates. \textbf{All results in this paper use 5.0}, i.e.\ the full pipeline. Setting it to 1.0 disables Stages~3--4 altogether and retains \(K\) directly; Table~\ref{tab:ratio1} shows that mode is statistically indistinguishable on standardized benchmarks, which is why we report the more expensive configuration here. \\
		\midrule
		prior\_strength & 0.3 & Stage 3 & Fusion ratio between dataset prior and adaptive score; constrains adaptive weight fluctuation and improves cross-scenario stability. \\
		\midrule
		n\_clusters & 0 & Stage 4 & Number of clustering constraints after fusion scoring. 0 disables clustering, selecting Top-\(K\) directly by composite score for stable industrial scenarios. \\
		\midrule
		manual\_weights & None & Stage 3/4 & Manually specified five-plugin fusion weights. If not None, overrides adaptive weights for on-demand plugin priority customization. \\
		\bottomrule
	\end{tabular*}
\end{table}

\section{Chunked Retrieval Before and After the Optimisation}\label{sec:app-chunk}

The optimisation of Section~\ref{sec:parallel-retrieval} was motivated by a concrete failure. Computing
per-patch cosine similarity against the memory bank materialises a $[C,M,D]$ tensor, and on MVTec-bottle
at $k{=}10$ that already demands 30.1~GB of inference memory at \texttt{chunk\_size}=256 and OOMs at
512---the size the un-chunked computation would need. Chunking the query patches was the workaround: it
bounds the intermediate tensor, but leaves memory and total latency tied to the chunking scheme, and the
groups are processed serially.

Table~\ref{tab:retrieval_chunk} reports the same configuration before and after the change, over five
chunk sizes.

\begin{table}
	\caption{Memory-bank retrieval on MVTec-bottle at $k{=}10$ (63 test images), before and after
		replacing per-patch cosine similarity with the normalised matrix product. Every cell is
		\emph{before~/~after}. P-AUPR is 0.7909 in every configuration. Chunking was only ever a
		workaround for the memory blow-up: the optimised path fits in memory at any chunk size, so the
		released code drops it altogether and the residual spread in the \emph{after} values is the
		serial-processing overhead of small chunks, not a memory constraint.}
	\label{tab:retrieval_chunk}
	\setlength{\tabcolsep}{4pt}
	\centering
	\scriptsize
	\begin{tabular*}{\tblwidth}{@{}LLLL@{}}
		\toprule
		chunk size & category time (s) & per image (s) & inference memory (GB) \\
		\midrule
		1   & 65.87\,/\,21.06 & 1.046\,/\,0.334 &  0.22\,/\,0.30 \\
		2   & 76.45\,/\,14.96 & 1.213\,/\,0.238 &  0.33\,/\,0.30 \\
		64  & 82.60\,/\, 8.03 & 1.311\,/\,0.127 &  7.61\,/\,0.30 \\
		128 & 81.93\,/\, 7.84 & 1.301\,/\,0.125 & 15.12\,/\,0.30 \\
		256 & 81.58\,/\, 9.42 & 1.295\,/\,0.150 & 30.14\,/\,0.30 \\
		512 & OOM\,/\, 9.08 & OOM\,/\,0.144 & OOM\,/\,0.30 \\
		\bottomrule
	\end{tabular*}
\end{table}

With the optimised path the inference memory is flat at 0.30~GB for every chunk size while category time
falls from 65.9--82.6~s to 7.8--21.1~s, a $4$--$10\times$ reduction. Because nothing OOMs any more, the
released code drops chunking entirely; the variation that remains across chunk sizes is the cost of
processing groups serially.

\section{Few-Shot Results}\label{sec:app-fewshot}

Table~\ref{tab:fewshot} reports the same five benchmarks under few-shot matching; it is referred to from Section~\ref{sec:arc-main-results} and is placed here for completeness.

\begin{table}
	\caption{Few-shot results on five benchmarks. Values are image-level/pixel-level. Best values are marked in red, second-best values are marked in blue.}
	\label{tab:fewshot}
	\setlength{\tabcolsep}{2pt}
	\centering
	\scriptsize 
	\begin{tabular*}{\tblwidth}{@{}LLLLLLL@{}}
		\toprule
		Datasets & Shots & Metrics & UniADet (CLIP) & UniADet (DINOv3) & FUSE (CLIP) & FUSE (DINOv3) \\
		\midrule
		\multirow{9}{*}{MVTec} & \multirow{3}{*}{1} & AUROC & \second{93.2}/94.8 & \best{93.5}/\best{96.6} & 91.9/92.0 & 93.1/\second{96.2} \\
		& & AUPR & \second{96.9}/52.1 & \best{97.1}/\second{56.9} & 95.8/48.7 & 96.3/\best{57.1} \\
		& & F1max & 92.8/53.0 & \second{93.0}/\best{56.6} & \best{94.0}/52.0 & 92.3/\second{55.6} \\
		\cmidrule{2-7}
		& \multirow{3}{*}{2} & AUROC & 93.4/95.2 & \second{93.8}/\best{96.9} & \best{95.0}/92.3 & 93.3/\second{96.5} \\
		& & AUPR & \second{97.1}/\second{53.5} & \best{97.3}/\best{57.6} & 97.0/52.0 & 96.4/\best{57.6} \\
		& & F1max & 92.9/54.5 & \second{93.4}/\best{57.4} & \best{94.7}/54.3 & 92.4/\second{55.9} \\
		\cmidrule{2-7}
		& \multirow{3}{*}{4} & AUROC & 93.5/95.7 & \second{94.0}/\best{97.1} & \best{96.7}/92.7 & 93.7/\second{96.7} \\
		& & AUPR & 97.1/54.8 & \second{97.5}/\best{58.3} & \best{98.2}/\second{55.1} & 96.6/\best{58.3} \\
		& & F1max & 93.0/55.5 & \second{93.5}/\best{58.2} & \best{95.8}/\second{57.1} & 92.7/56.7 \\
		\midrule
		\multirow{9}{*}{VisA} & \multirow{3}{*}{1} & AUROC & \second{93.0}/\second{96.5} & \best{96.2}/\best{97.3} & 85.6/96.2 & 92.3/\best{97.3} \\
		& & AUPR & \second{75.5}/26.6 & \best{86.2}/\best{30.2} & 54.0/24.1 & 74.4/\second{27.1} \\
		& & F1max & 71.2/\second{33.3} & \best{82.4}/\best{36.1} & 56.7/32.3 & \second{72.1}/\second{33.3} \\
		\cmidrule{2-7}
		& \multirow{3}{*}{2} & AUROC & \second{93.5}/96.6 & \best{96.4}/\best{97.4} & 88.8/96.4 & 92.6/\best{97.4} \\
		& & AUPR & \second{77.2}/27.5 & \best{86.9}/\best{30.6} & 60.5/26.6 & 75.5/\second{27.6} \\
		& & F1max & 73.1/34.3 & \best{83.4}/\best{36.4} & 61.7/\second{34.7} & \second{73.1}/33.8 \\
		\cmidrule{2-7}
		& \multirow{3}{*}{4} & AUROC & \second{93.8}/96.8 & \best{96.4}/\best{97.5} & 90.9/\second{96.8} & 92.7/\best{97.5} \\
		& & AUPR & \second{78.2}/28.0 & \best{87.0}/\best{30.6} & 66.0/\second{29.8} & 75.7/27.6 \\
		& & F1max & 73.6/34.8 & \best{83.7}/\second{36.6} & 66.1/\best{38.2} & \second{73.7}/33.8 \\
		\midrule
		\multirow{9}{*}{BTAD} & \multirow{3}{*}{1} & AUROC & 93.5/95.7 & \second{94.7}/\best{98.2} & 92.0/90.1 & \best{95.5}/\second{97.4} \\
		& & AUPR & \best{96.4}/54.1 & \best{96.4}/\best{61.2} & 88.3/46.2 & \second{95.2}/\second{58.6} \\
		& & F1max & 92.5/56.1 & \best{94.6}/\best{57.9} & 86.5/50.1 & \second{93.5}/\second{56.2} \\
		\cmidrule{2-7}
		& \multirow{3}{*}{2} & AUROC & 93.9/96.0 & \second{94.8}/\best{98.2} & 93.9/90.5 & \best{95.6}/\second{97.4} \\
		& & AUPR & \best{96.8}/56.5 & \second{96.2}/\best{61.5} & 95.2/50.0 & 95.4/\second{59.0} \\
		& & F1max & 93.3/\second{57.6} & \best{94.6}/\best{58.2} & 92.3/52.6 & \second{93.5}/56.4 \\
		\cmidrule{2-7}
		& \multirow{3}{*}{4} & AUROC & 94.1/96.3 & 94.9/\best{98.3} & \second{94.9}/90.8 & \best{95.8}/\second{97.5} \\
		& & AUPR & \best{96.7}/56.8 & \second{96.4}/\best{61.7} & 95.9/50.6 & 95.6/\second{59.3} \\
		& & F1max & 92.4/\second{57.9} & \best{95.1}/\best{58.5} & 92.4/53.0 & \second{93.5}/56.6 \\
		\midrule
		\multirow{9}{*}{KSDD} & \multirow{3}{*}{1} & AUROC & \second{95.3}/97.3 & 86.0/\best{98.8} & \best{98.2}/95.8 & 85.0/\best{98.8} \\
		& & AUPR & \second{75.3}/\second{22.5} & 48.2/15.6 & \best{89.8}/\best{39.3} & 44.3/20.6 \\
		& & F1max & \second{77.7}/\second{38.3} & 54.6/26.5 & \best{81.3}/\best{48.4} & 51.0/32.0 \\
		\cmidrule{2-7}
		& \multirow{3}{*}{2} & AUROC & \second{95.2}/97.3 & 86.2/\best{98.7} & \best{97.2}/95.9 & 84.8/\best{98.7} \\
		& & AUPR & \second{75.1}/\second{22.9} & 48.3/15.6 & \best{89.4}/\best{42.6} & 43.8/20.6 \\
		& & F1max & \second{75.0}/\second{38.7} & 55.7/26.6 & \best{83.5}/\best{50.9} & 50.9/31.9 \\
		\cmidrule{2-7}
		& \multirow{3}{*}{4} & AUROC & \second{95.1}/97.4 & 86.5/\second{98.7} & \best{97.1}/95.9 & 85.3/\best{98.8} \\
		& & AUPR & \second{75.4}/\second{23.8} & 49.2/16.0 & \best{90.8}/\best{44.0} & 44.9/21.1 \\
		& & F1max & \second{74.7}/\second{39.8} & 55.7/27.1 & \best{87.6}/\best{52.1} & 52.1/32.4 \\
		\midrule
		\multirow{9}{*}{Real-IAD} & \multirow{3}{*}{1} & AUROC & 76.7/96.4 & \best{79.3}/\best{97.2} & 67.2/94.3 & \second{78.5}/\best{97.2} \\
		& & AUPR & 84.8/37.7 & \best{87.3}/\best{43.1} & 79.6/19.1 & \second{87.0}/\second{41.9} \\
		& & F1max & 82.2/41.7 & \best{82.9}/\best{45.4} & 80.2/25.6 & \second{82.3}/\second{44.4} \\
		\cmidrule{2-7}
		& \multirow{3}{*}{2} & AUROC & 77.7/96.8 & \best{79.9}/\best{97.5} & 70.6/94.9 & \second{79.1}/\best{97.5} \\
		& & AUPR & 85.4/38.8 & \best{87.7}/\best{44.0} & 80.9/21.2 & \second{87.6}/\second{43.0} \\
		& & F1max & \second{82.5}/42.3 & \best{83.1}/\best{46.0} & 81.1/27.1 & 82.4/\second{45.0} \\
		\cmidrule{2-7}
		& \multirow{3}{*}{4} & AUROC & 78.3/97.2 & \best{80.3}/\best{97.9} & 73.7/95.5 & \second{79.5}/\second{97.8} \\
		& & AUPR & 85.8/39.9 & \best{88.1}/\best{44.8} & 82.5/24.2 & \second{88.0}/\second{43.6} \\
		& & F1max & 82.5/43.2 & \best{83.2}/\best{46.4} & 81.7/30.7 & \second{82.7}/\second{45.4} \\
		\bottomrule
	\end{tabular*}
\end{table}

\section{Qualitative Comparison}\label{sec:app-vis}

\begin{figure}
	\centerline{\includegraphics[width=\columnwidth]{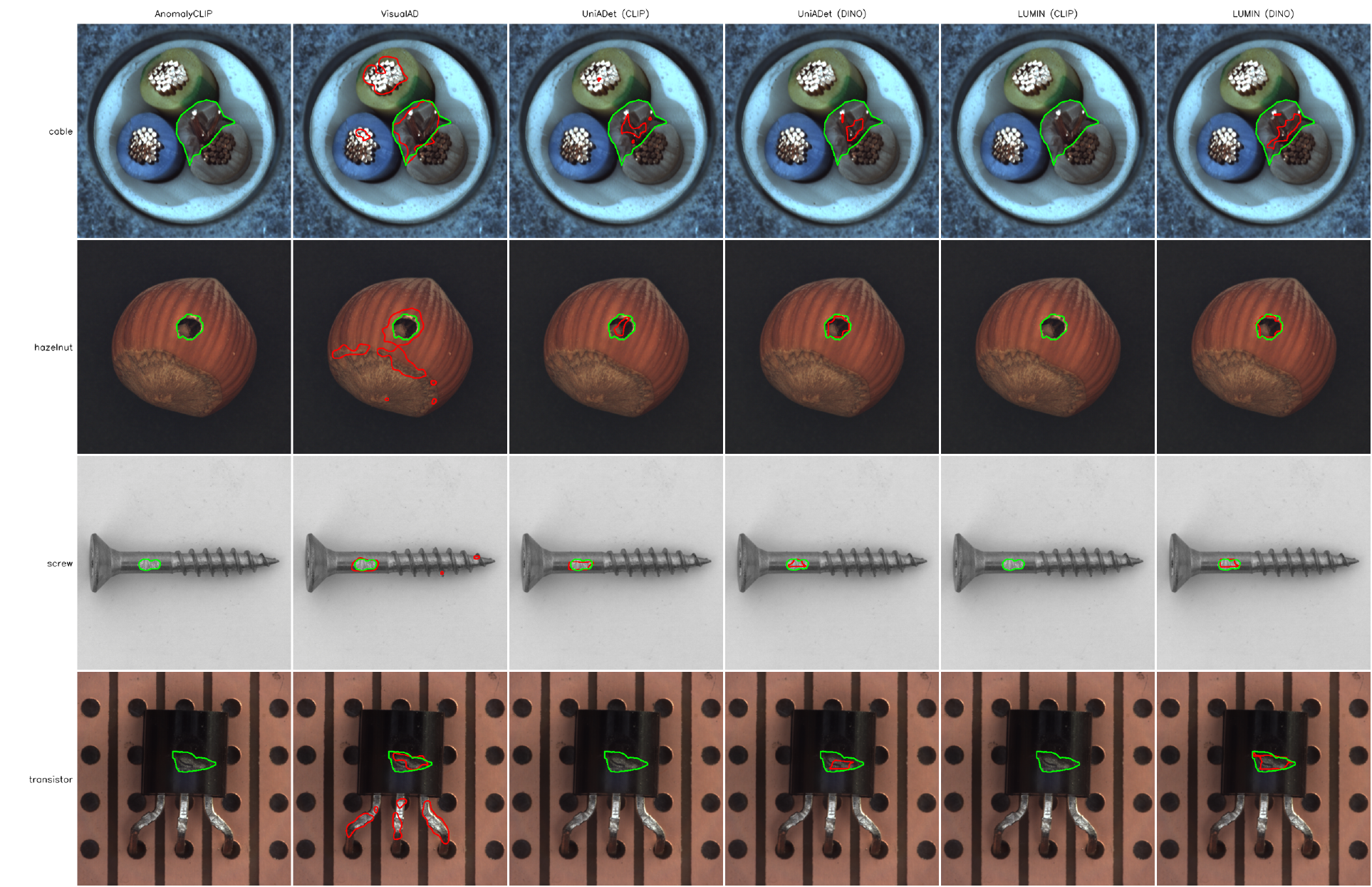}}
	\caption{Segmentation visualization comparison across methods on cable, hazelnut, screw, and transistor. Green: ground truth; red: predictions.}
	\label{fig:vis}
\end{figure}

We select four representative industrial defect categories---cable, hazelnut, screw, and transistor---and compare the segmentation visualizations of AnomalyCLIP, VisualAD, UniADet (CLIP), UniADet (DINOv3), FUSE (CLIP), and FUSE (DINOv3), as shown in Fig.~\ref{fig:vis}.

The baselines fail in different ways rather than uniformly. VisualAD suffers from severe global false positives, generating large spurious anomaly regions on defect-free colored components (cable), normal hazelnut shell bottoms, and intact transistor pins. It can only roughly localize the defect main body, failing to distinguish normal textures from damaged regions, with severely distorted defect boundaries. AnomalyCLIP localizes the defect body closely on hazelnut, screw, and transistor, with contours that track the ground truth; on cable, however, it additionally fires on an intact component, and its boundaries are coarser than the ground-truth outline throughout. The CLIP and DINOv3 variants of UniADet can localize the defect main body, but both suffer from local prediction shifts and fragmented false positives. UniADet (CLIP) shows boundary misalignment on cable and transistor, with unnecessary expansion on screw defects; UniADet (DINOv3) alleviates these shifts but still generates scattered false predictions around hazelnut hole edges and transistor shells, indicating inherent limitations in local feature alignment.

Among our FUSE variants, FUSE (CLIP) exhibits clear localization failure, producing almost no valid anomaly contours across all four categories, failing to identify defective regions. In contrast, FUSE (DINOv3)---with the DINO backbone compensating for CLIP's feature-matching deficiencies---is the only configuration that localizes all four categories without spurious responses: it eliminates the global and local false positives common in the baselines, producing no spurious predictions on defect-free regions (e.g., mixed-component cable, intact transistor pins, normal shells, and threaded areas) with a clear normal-anomaly boundary, and its predictions closely match ground truth across varying defect scales and morphologies. AnomalyCLIP is the closest competitor but retains a false positive on cable and coarser boundaries; the remaining methods are clearly below both.

\section{Exploration Paths: The Roles of Single Branch and FUSE}\label{sec:app-path}

This appendix reviews the two parameter-reduction paths we explored on UniADet (Double Branches), clarifies their roles and boundary conditions, and reports FUSE's failure on CLIP as an empirical reference.

\textbf{Path One: Single Branch---initial lightweighting.} Single Branch investigates whether a dedicated classification head is indispensable for image-level anomaly scoring. Our findings indicate that it is not. Removing the classification branch alone halves the number of trainable parameters, while preserving pixel-wise detection performance on VisA, BTAD, and Real-IAD. For production-line scenarios that demand localization only, Single Branch thus occupies a Pareto-optimal position. For applications requiring image-level judgment, we resort to Top-K average pooling over the pixel anomaly map as a proxy, which yields reliable estimates at the cost of only a marginal few-point drop in I-AUROC.

Notably, we incorporate this Top-K pooling solely during inference, rather than using it as an auxiliary training objective. While integrating image-level supervision during training could potentially boost overall performance, it would inevitably introduce extra architectural complexity and computational overhead. Since our primary focus is structural lightweighting, we deliberately exclude such auxiliary branches and instead treat the inference-only Top-K strategy as a pragmatic, parameter-free reference for image-level discriminability.

\begin{table}
	\caption{Mean results over five benchmarks. Values are of FUSE/UniADet. 2L, 4L and 5L means training and inference with features (from vision backbone layers) $\{12,24\}$, $\{6,12,18,24\}$ and $\{12,15,18,21,24\}$. Best values are marked in red, second-best values are marked in blue.}
	\label{tab:layer_compare}
	\setlength{\tabcolsep}{4pt}
	\centering
	\scriptsize
	\begin{tabular*}{\tblwidth}{@{}LLLLLLLL@{}}
		\toprule
		$k$ & Metric & CLIP 2L & CLIP 4L & CLIP 5L & DINOv3 2L & DINOv3 4L & DINOv3 5L \\
		\midrule
		0 & I-AUROC & 53.7 / 85.4 & 54.4 / \second{87.0} & 51.9 / \best{87.6} & \second{81.6} / 86.0 & \best{82.2} / 85.0 & 81.4 / 86.2 \\
		0 & I-AUPR & 51.3 / \best{82.6} & 51.2 / \best{82.6} & 49.9 / \second{82.5} & \second{71.1} / 81.1 & \best{71.6} / 76.9 & 70.7 / 77.6 \\
		0 & I-F1max & 58.5 / \second{80.3} & 58.3 / \best{80.9} & 57.6 / \second{80.3} & \second{71.0} / 78.2 & \best{71.6} / 76.8 & \second{71.0} / 77.2 \\
		0 & P-AUROC & 79.8 / 94.5 & 79.1 / 94.3 & 78.4 / 94.7 & \best{94.9} / \second{95.4} & \second{94.5} / 95.3 & \best{94.9} / \best{96.0} \\
		0 & P-AUPR & 4.4 / 31.8 & 4.2 / 32.0 & 4.0 / \best{32.9} & \second{34.5} / \second{35.9} & \best{34.8} / 35.5 & \second{34.5} / \best{37.8} \\
		0 & P-F1max & 8.2 / 36.3 & 7.6 / \second{36.8} & 7.5 / \best{37.5} & 38.5 / \second{39.3} & \best{39.0} / 39.1 & \second{38.6} / \best{41.1} \\
		1 & I-AUROC & 85.9 / 90.3 & 87.0 / 90.3 & 86.3 / \second{90.7} & 88.7 / 89.5 & \second{88.9} / 90.0 & \best{90.0} / \best{92.1} \\
		1 & I-AUPR & 80.9 / \best{87.0} & \best{81.5} / 85.8 & 80.0 / \best{87.0} & 79.1 / 83.6 & 79.4 / 83.0 & \second{81.1} / \second{86.0} \\
		1 & I-F1max & \best{80.4} / 83.6 & 79.7 / 83.3 & 79.5 / \second{84.2} & 78.0 / 81.1 & 78.2 / 81.5 & \second{79.8} / \best{85.0} \\
		1 & P-AUROC & 93.1 / 96.0 & 93.7 / 96.1 & 92.6 / 95.8 & \second{97.3} / 97.5 & \second{97.4} / \second{97.6} & \best{97.5} / \best{97.7} \\
		1 & P-AUPR & 32.8 / 37.6 & 35.5 / 38.6 & 31.5 / 38.2 & 40.5 / \second{41.4} & \second{41.1} / \second{41.4} & \best{42.1} / \best{44.1} \\
		1 & P-F1max & 40.2 / 43.4 & 41.7 / \second{44.5} & 39.0 / 43.7 & 43.8 / \second{44.5} & \second{44.3} / \second{44.5} & \best{45.3} / \best{46.8} \\
		2 & I-AUROC & 88.1 / 90.8 & \second{89.1} / 90.7 & 88.5 / \second{91.2} & 89.0 / 89.8 & \second{89.1} / 90.2 & \best{90.2} / \best{92.4} \\
		2 & I-AUPR & 83.7 / \second{87.4} & \best{84.6} / 86.3 & \second{83.6} / \best{87.6} & 79.6 / 84.1 & 79.7 / 83.3 & 81.5 / 86.4 \\
		2 & I-F1max & \second{82.5} / 84.1 & \best{82.6} / 83.4 & 82.4 / \second{84.8} & 78.1 / 81.4 & 78.5 / 82.0 & 80.3 / \best{85.2} \\
		2 & P-AUROC & 93.5 / 96.2 & 94.0 / \second{96.4} & 92.9 / 96.0 & 97.4 / 97.6 & \second{97.5} / \second{97.7} & \best{97.6} / \best{97.8} \\
		2 & P-AUPR & 35.5 / 38.6 & 38.5 / 39.8 & 34.7 / 39.3 & 41.1 / \second{41.9} & \second{41.6} / 41.8 & \best{42.7} / \best{44.7} \\
		2 & P-F1max & 42.2 / 44.3 & 43.9 / \second{45.5} & 41.3 / 44.6 & 44.1 / 44.8 & \second{44.6} / 44.9 & \best{45.7} / \best{47.2} \\
		4 & I-AUROC & 89.8 / 91.0 & \best{90.7} / 91.0 & 90.0 / \second{91.4} & 89.3 / 90.1 & 89.4 / 90.4 & \second{90.5} / \best{92.6} \\
		4 & I-AUPR & \second{86.0} / \best{87.8} & \best{86.7} / 86.6 & 85.3 / \best{87.8} & 80.1 / 84.5 & 80.1 / 83.6 & 81.7 / 86.7 \\
		4 & I-F1max & \second{84.0} / 84.4 & \best{84.7} / 83.3 & 83.7 / \second{84.7} & 78.6 / 82.0 & 78.9 / 82.2 & 80.6 / \best{85.5} \\
		4 & P-AUROC & 93.8 / 96.5 & 94.3 / 96.7 & 93.3 / 96.3 & 97.6 / 97.8 & \second{97.7} / \second{97.9} & \best{97.8} / \best{98.0} \\
		4 & P-AUPR & 37.7 / 39.5 & 40.7 / 40.7 & 36.4 / 40.1 & 41.5 / \second{42.3} & \second{42.0} / \second{42.3} & \best{43.2} / \best{45.2} \\
		4 & P-F1max & 44.6 / 45.1 & \best{46.2} / \second{46.2} & 43.2 / 45.3 & 44.5 / 45.2 & \second{45.0} / 45.3 & \best{46.2} / \best{47.7} \\
		\bottomrule
	\end{tabular*}
\end{table}

\textbf{Path Two: FUSE---extreme lightweighting.} FUSE further compresses multiple segmentation heads into a single set and introduces cross-layer self-attention fusion (FeatFusion) before detection. On DINOv3, pixel-level metrics decline minimally, as shown in Table~\ref{tab:layer_compare}.

\begin{figure}
	\centering
	\includegraphics[width=\linewidth]{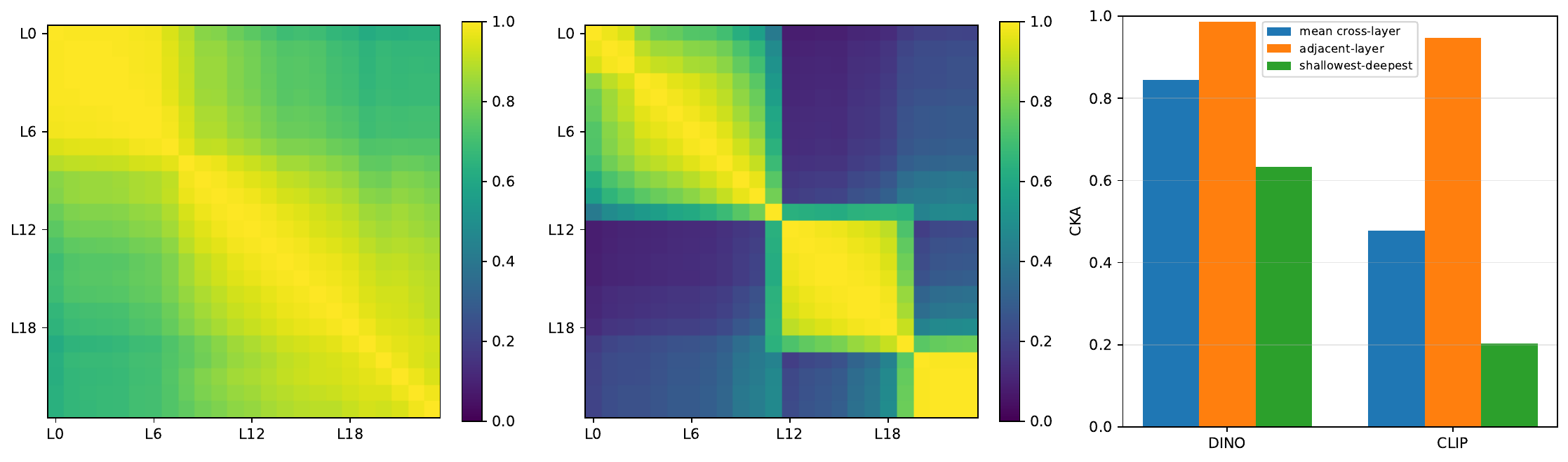}
	\caption{\textbf{Cross-layer representation consistency.} Left: inter-layer linear
		CKA~\cite{cka} of DINOv3 ViT-L/16 across layers 0--23. Middle: the same for CLIP ViT-L/14.
		Right: three CKA statistics (mean cross-layer, adjacent-layer, and shallowest--deepest) for
		both backbones. Brighter is more similar. DINOv3 is uniformly consistent (mean $0.845$);
		CLIP decomposes into blocks of internally consistent layers (0--9, 10--17, 18--23) with
		near-zero cross-block similarity (mean $0.477$, $0.203$ between the extreme layers).
		Adjacent-layer similarity is high for both ($0.987$ vs.\ $0.947$), so the two differ in how
		\emph{far} consistency extends, not in whether it exists locally.}
	\label{fig:cka}
\end{figure}

However, FUSE's effectiveness depends critically on backbone inter-layer consistency. In the \textbf{zero-shot setting ($k{=}0$) FUSE fails completely on CLIP}, producing near-random pixel-level scores (Table~\ref{tab:0shot}, MVTec P-AUPR 12.2 against UniADet-CLIP's 43.5); once a memory bank is supplied the gap narrows substantially, so the failure is a property of the zero-shot regime, not of the fusion design in general. Our hypothesis is that CLIP's strong inter-layer heterogeneity (shallow texture vs.\ deep semantics) cannot be preserved by single-head fusion, whereas DINOv3's self-supervised features are highly consistent across layers. Table~\ref{tab:layer_compare} is consistent with this reading: increasing the number of sampled layers improves DINOv3 (P-F1max 38.5$\to$39.0) but leaves CLIP essentially unchanged (P-AUPR 4.4$\to$4.0). \textbf{The premise is measurable; the causal step is not.} Fig.~\ref{fig:cka} reports inter-layer linear CKA~\cite{cka} for both backbones: DINOv3's matrix is uniformly bright (mean $0.845$), whereas CLIP's has a clear block structure---layers 0--9, 10--17 and 18--23 each internally consistent, with cross-block similarity near zero (mean $0.477$, $0.203$ between the extreme layers). Adjacent-layer similarity is high for both ($0.987$ vs.\ $0.947$), so the two backbones differ in how \emph{far} consistency extends, not in whether it exists locally. This establishes the premise---CLIP's layers really are far more heterogeneous across depth---but not the mechanism. The step from that heterogeneity to the observed single-head fusion failure remains an inference drawn from two backbones, and Table~\ref{tab:layer_compare} is not even monotone ($k{=}0$ P-F1max reads 38.5, 39.0, 38.6 across the three layer settings), so we report the correspondence as supporting evidence and do not call it the root cause.

\textbf{Outlook: from global early fusion to local feature enhancement.} While FUSE's fusion causes information loss, its degradation on DINOv3 is limited and justifies further optimization. A natural improvement is to keep multi-layer independent heads and shift fusion granularity downward---provide each head with locally enhanced features from neighboring layers (e.g., layer~12 aggregates layers~$\{9,12,15\}$), rather than completing full fusion before detection. This injects inter-layer neighborhood information while preserving each layer's independent processing capability, offering a finer balance between information preservation and parameter efficiency. However, storage overhead grows linearly with layer count, conflicting with the lightweighting theme of this paper; we leave this trade-off as future work.

\printcredits

\bibliographystyle{cas-model2-names}

\bibliography{cas-refs}

\end{document}